\documentclass[10pt,twocolumn,letterpaper]{article}

\usepackage[final,datasets]{wacv}      

\usepackage{booktabs}
\usepackage{adjustbox}
\usepackage{multirow}
\usepackage{colortbl}
\usepackage{pifont}
\usepackage{xcolor}

\definecolor{dmgreen1}{HTML}{519D78}
\definecolor{dmgreen2}{HTML}{8BCD8B}
\definecolor{dmgreen3}{HTML}{AADCA9}
\definecolor{dmgreen4}{HTML}{C4E9CA}
\definecolor{dmgreen5}{HTML}{DDF3DE}
\definecolor{dmgreen6}{HTML}{F3FBF2}
\definecolor{dmblue1}{HTML}{5385BD}
\definecolor{dmblue2}{HTML}{9BC7DF}
\definecolor{dmblue3}{HTML}{DFF3F8}

\newcommand{\DMtotalHours}{393}
\newcommand{\DMtotalSeqs}{4,367}
\newcommand{\DMtotalWindows}{680,082}
\newcommand{\DMtotalDrivers}{360}
\newcommand{\DMbatonRoutes}{1,329}
\newcommand{\DMbatonHours}{319}
\newcommand{\DMbatonDrivers}{301}

\newcommand{\DMwebHours}{71.7}
\newcommand{\DMwebDrivers}{58}
\newcommand{\DMwebCreators}{25}
\newcommand{\DMwebAnnotClips}{1,868}
\newcommand{\DMwebDriverClips}{873}

\newcommand{\DMaideClips}{2,898}
\newcommand{\DMaideHours}{2.4}
\newcommand{\DMwebViewBack}{661}
\newcommand{\DMwebViewSide}{3147}
\newcommand{\DMwebViewFront}{1050}

\newcommand{\std}[1]{{\scriptsize$\pm$#1}}

\definecolor{wacvblue}{rgb}{0.21,0.49,0.74}
\usepackage[pagebackref,breaklinks,colorlinks,allcolors=wacvblue]{hyperref}

\def\wacvPaperID{3041}
\def\confName{WACV}
\def\confYear{2027}

\title{DriveMotion: A Large-Scale Multi-Source Benchmark for\\Driver Motion Sequence Modeling and Forecasting}

\author{Yuhang Wang\textsuperscript{1}, \; Chuheng Wei\textsuperscript{2}, \; Jingxin Yang\textsuperscript{3}, \; Xishun Liao\textsuperscript{4}, \; Hao Zhou\textsuperscript{1}\\[4pt]
{\normalsize \textsuperscript{1}University of South Florida \quad \textsuperscript{2}Purdue University \quad \textsuperscript{3}Stanford University \quad \textsuperscript{4}University of Central Florida}\\[4pt]
{\normalsize \href{https://wangyuhang-cmd.github.io/drivemotion/}{\texttt{wangyuhang-cmd.github.io/drivemotion}}}\\[2pt]
{\small \texttt{\{yuhangw, haozhou1\}@usf.edu \; wch@purdue.edu \; jingxiny@stanford.edu \; xishun.liao@ucf.edu}}
}

\begin{document}
\maketitle
\begin{abstract}
Driver motion can provide cues to ongoing behavior, attention, and near-term driving intent.
However, most existing driver-centric datasets focus on recognizing predefined driver behaviors from short video clips, while human motion forecasting benchmarks largely target motion outside the vehicle. 
We introduce \textbf{DriveMotion}, a multi-source benchmark for continuous driver motion forecasting. 
DriveMotion contains \DMtotalHours{} hours of 133-keypoint motion sequences at 10\,Hz from \DMtotalDrivers{} drivers, integrating naturalistic driving data, curated public in-cabin videos, and the AIDE dataset into a unified representation with per-joint validity masks and synchronized driving context. 
Naturalistic driving contains long periods of limited body movement, making uniformly sampled evaluation dominated by persistence and less sensitive to brief but behaviorally meaningful motion. 
To address this, we use dynamics-anchored evaluation, placing forecasting windows around vehicle maneuvers identified offline from CAN signals without providing CAN to the model at inference.
Arm motion in pre-maneuver windows is $3.4\times$ greater than in route-matched stable-driving controls. On these anchored windows, learned models reduce forecasting error over persistence by up to 15\%, while maneuver-enriched training improves forecast-derived Part-State F1 by 44\% over the zero-motion reference.
Training on the full multi-source corpus further reduces forecasting error on held-out web drivers by 38\% compared with BATON-only training. 
DriveMotion provides identity-disjoint splits, fixed evaluation subsets, and reference implementations for reproducible evaluation of continuous driver motion forecasting.
\end{abstract}

\section{Introduction}
\label{sec:intro}

\begin{figure*}[htbp]
  \centering
  \includegraphics[width=0.9\linewidth]{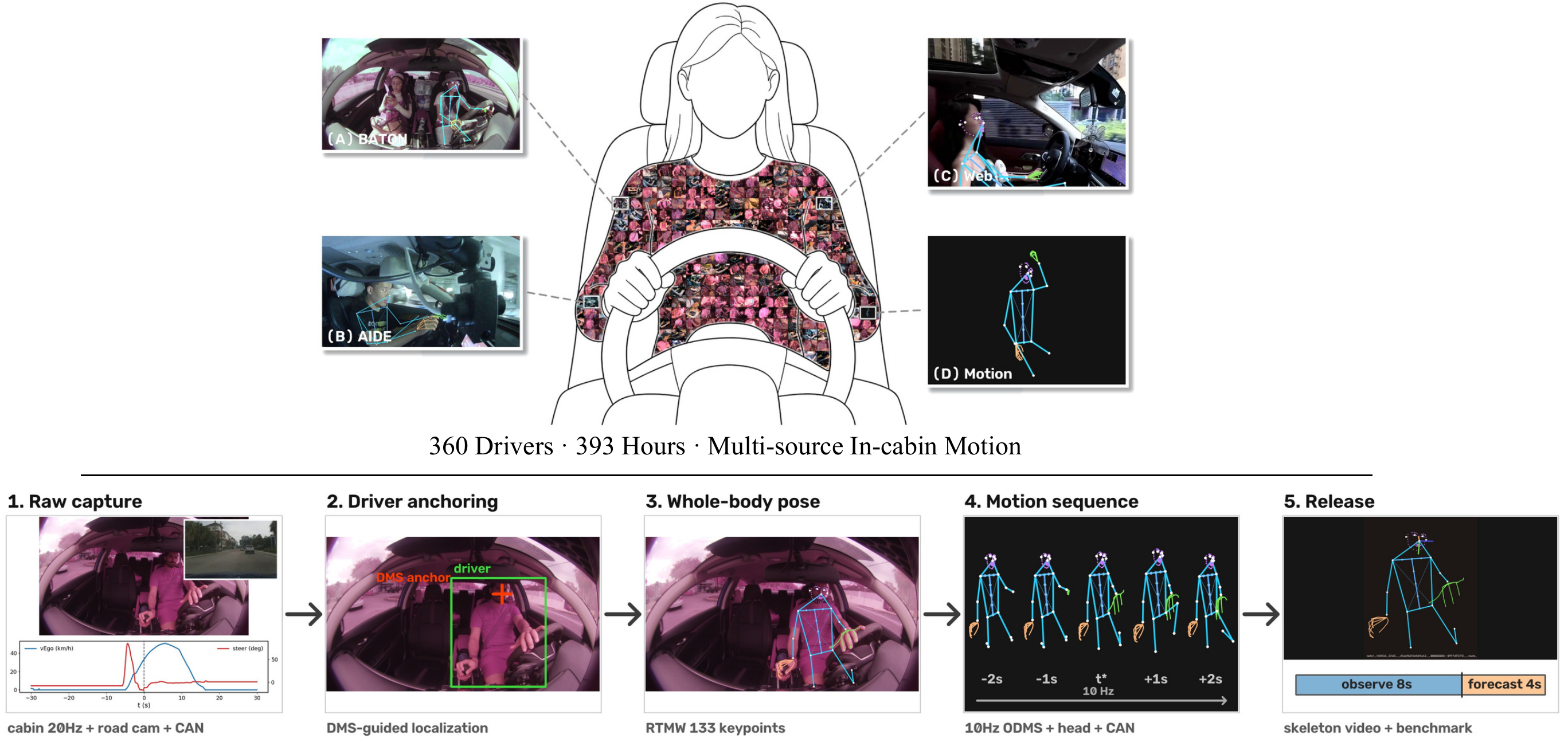}
  \caption{Overview of DriveMotion and its benchmark construction pipeline.}
  \label{fig:teaser}
\end{figure*}

Driver motion can provide cues to ongoing behavior, visual attention, and near-term maneuver intent~\cite{martin2019driveandact,jain2021mdmd,jain2016brain4cars,ma2023cemformer}. A driver may turn toward a mirror before changing lanes, reach toward the center console, or shift posture before taking over vehicle control. These movements unfold over several seconds and are relevant to takeover readiness and distraction monitoring~\cite{jain2016brain4cars,ma2023cemformer,riskaware2026}. 
Beyond recognizing driver actions after they occur, forecasting aims to predict how driver motion evolves over the next several seconds.

Existing datasets are not well suited for this task. Driver-centric datasets such as Drive\&Act, DMD, 100-Driver, and DAD~\cite{martin2019driveandact,ortega2020dmd,wang2023hundreddriver,kopuklu2021dad,jain2021mdmd,nobari2024mand,daos2026,avdm2024} primarily formulate driver understanding as recognition from short clips with limited temporal continuity. Human motion forecasting has mature sequence benchmarks~\cite{martinez2017human,mao2019learning,guo2023simlpe,barquero2023belfusion,chen2023humanmac}, but these benchmarks are largely based on laboratory capture~\cite{ionescu2014human36m,mahmood2019amass} or outdoor pedestrians~\cite{vonmarcard20183dpw,adeli2021tripod,waymo3dskelmo2025,vcpose2026}. However, in-cabin motion is different. Drivers are seated, large parts of the body may be occluded, camera placement varies across recordings, and long periods contain little body movement. Continuous driver motion forecasting in realistic cabin settings therefore remains largely unaddressed.

Building such benchmark raises several challenges. 
Long, continuous in-cabin motion data are difficult to collect, and most existing datasets are fixed collections that are not designed to incorporate new recordings or additional sources over time. 
Forecasting requires temporally continuous recordings rather than independently labeled clips, while in-cabin video varies substantially in camera placement and visible body regions. 
Partial observability must therefore be represented explicitly rather than filled by imputation. 
During naturalistic driving, especially in stable-driving periods, drivers often remain nearly still for long stretches. 
Uniform sampling therefore selects many near-static windows and misses the brief movements that are more informative for forecasting.

We introduce \textbf{DriveMotion}, a benchmark for continuous driver motion forecasting (\cref{fig:teaser}). DriveMotion contains \DMtotalHours{} hours of motion from \DMtotalDrivers{} drivers in a common 133-keypoint representation at 10\,Hz. 
It combines three sources with different roles.
BATON~\cite{baton2026} provides long naturalistic in-cabin recordings with synchronized CAN.
Our curated collection of publicly available in-cabin videos adds variation in camera placement and body visibility.
AIDE~\cite{yang2023aide} provides behavior and emotion annotations.
All three sources are processed using the same extraction and quality-control pipeline with per-joint validity masks.
To reduce the effect of persistence-dominated evaluation, DriveMotion also anchors evaluation windows to vehicle maneuvers identified offline from CAN signals. 
Identity-disjoint splits and optional synchronized exterior scene features are provided for benchmark evaluation.

DriveMotion makes three main contributions. 
First, it provides \DMtotalHours{} hours of multi-source in-cabin motion from \DMtotalDrivers{} drivers in a unified 133-keypoint representation, combining long naturalistic sequences, variation in camera placement and body visibility, synchronized driving context, and semantic annotations. 
Second, it establishes a driver motion forecasting benchmark with identity-disjoint splits and evaluation over pre-maneuver, post-maneuver, and matched stable-driving intervals, using vehicle dynamics only offline to construct the evaluation windows. 
Third, our experiments show that forecasting performance is concentrated in motion-active, maneuver-related intervals: arm motion near maneuvers is $3.4\times$ that of matched stable driving, learned models reduce forecasting error over persistence by up to 15\%, maneuver-enriched training improves forecast-derived Part-State F1 by 44\% over the zero-motion reference, and training on the full multi-source corpus reduces error on held-out web drivers by 38\% compared with BATON-only training.

\begin{table*}[t]
  \centering
  \footnotesize
  \setlength{\tabcolsep}{3.5pt}
  \caption{Comparison with driver-facing datasets. \emph{Ext.\ vid.}: synchronized exterior video; \emph{Dyn.\ ev.}: vehicle-dynamics events; \emph{View/vis.}: viewpoint and visibility labels; \emph{Sem.}: semantic labels.}
  \label{tab:comparison}
  \rowcolors{2}{dmblue3}{white}
  \begin{tabular}{lcccccccccccc}
    \toprule
    \rowcolor{dmblue2}
    Dataset & Hours & Drivers & Views & Motion & Cont. & CAN & Ext.\ vid. & Dyn.\ ev. & Head GT & Sem. & View/vis. & Forecast \\
    \midrule
    Drive\&Act~\cite{martin2019driveandact} & 12 & 15 & 6 & 3D pose & \ding{55} & \ding{55} & \ding{55} & \ding{55} & \ding{55} & \checkmark & \ding{55} & \ding{55} \\
    DMD~\cite{ortega2020dmd} & 41 & 37 & 3 & \ding{55} & \ding{55} & \ding{55} & \ding{55} & \ding{55} & \ding{55} & \checkmark & \ding{55} & \ding{55} \\
    MDMD~\cite{jain2021mdmd} & -- & 59 & 2 & \ding{55} & \ding{55} & \ding{55} & \ding{55} & \ding{55} & \checkmark & \checkmark & \ding{55} & \ding{55} \\
    HDD~\cite{ramanishka2018hdd} & 104 & -- & road & \ding{55} & \checkmark & \checkmark & \checkmark & \checkmark & \ding{55} & \checkmark & \ding{55} & \ding{55} \\
    100-Driver~\cite{wang2023hundreddriver} & -- & 100 & 4 & \ding{55} & \ding{55} & \ding{55} & \ding{55} & \ding{55} & \ding{55} & \checkmark & \ding{55} & \ding{55} \\
    DAD~\cite{kopuklu2021dad} & $\approx$11 & 31 & 2 & \ding{55} & \ding{55} & \ding{55} & \ding{55} & \ding{55} & \ding{55} & \checkmark & \ding{55} & \ding{55} \\
    DD-Pose~\cite{roth2019ddpose} & -- & 27 & 2 & \ding{55} & \ding{55} & \checkmark & \ding{55} & \ding{55} & \checkmark & \ding{55} & \ding{55} & \ding{55} \\
    manD 1.0~\cite{nobari2024mand} & -- & 50 & -- & \ding{55} & \ding{55} & \checkmark & -- & \checkmark & \ding{55} & \ding{55} & \ding{55} & \ding{55} \\
    DAOS~\cite{daos2026} & -- & -- & -- & \ding{55} & \ding{55} & \ding{55} & -- & -- & \ding{55} & -- & \ding{55} & \ding{55} \\
    AVDM~\cite{avdm2024} & -- & -- & -- & \ding{55} & \checkmark & \ding{55} & -- & -- & \ding{55} & -- & \ding{55} & \ding{55} \\
    AIDE~\cite{yang2023aide} & 2.4 & -- & 4 & 2D pose & \ding{55} & \ding{55} & \ding{55} & \ding{55} & \ding{55} & \checkmark & \ding{55} & \ding{55} \\
    \midrule
    \rowcolor{dmblue2}
    \textbf{DriveMotion (ours)} &
    \textbf{\DMtotalHours} &
    \textbf{\DMtotalDrivers} &
    \textbf{5 types} &
    \textbf{133-kpt seq.} &
    \checkmark &
    \checkmark &
    \checkmark &
    \checkmark &
    \checkmark &
    \checkmark &
    \checkmark &
    \checkmark \\
    \bottomrule
  \end{tabular}
\end{table*}

\section{Related Work}
\label{sec:related}

\paragraph{Driver-centered datasets.}
Existing driver-monitoring datasets, including Drive\&Act~\cite{martin2019driveandact}, DMD~\cite{ortega2020dmd}, 100-Driver~\cite{wang2023hundreddriver}, DAD~\cite{kopuklu2021dad}, MDMD~\cite{jain2021mdmd}, manD~1.0~\cite{nobari2024mand}, AVDM~\cite{avdm2024}, DAOS~\cite{daos2026}, DD-Pose~\cite{roth2019ddpose}, HDD~\cite{ramanishka2018hdd}, and AIDE~\cite{yang2023aide}, formulate driver understanding as recognition of activities, states, or anomalies from short labeled clips (Table~\ref{tab:comparison}). 
To our knowledge, none releases continuous whole-body motion sequences or a standardized forecasting protocol. 
BATON~\cite{baton2026} supplies the long naturalistic recordings behind our largest source;
DriveMotion converts them into quality-controlled motion sequences and integrates them into a unified benchmark.


\paragraph{Driver understanding and intent prediction.}
Driver-centered methods have progressed from sensory-fusion RNNs that anticipate maneuvers seconds ahead~\cite{jain2016brain4cars} to transformers that fuse in-cabin and exterior views for driver intent~\cite{ma2023cemformer}.
Pose-based models~\cite{peng2022transdarc,posevinet2023,poguise2024,sdatr2025} further show that body keypoints provide a compact representation of driver state and activity.
More recent systems extend prediction beyond discrete intent labels: Cockpit-Llama predicts driver intent from historical interactions and cockpit states~\cite{chen2025cockpitllama}, while Driver-WM models future in-cabin dynamics through traffic-conditioned latent rollouts~\cite{driverwm2026}.
These works motivate continuous driver-motion data that support prediction beyond short labeled clips.

\paragraph{Human motion forecasting.}
Human motion forecasting has mature sequence models, including recurrent and graph-based methods~\cite{martinez2017human,yan2018stgcn,mao2019learning,mao2020history}, MLP baselines~\cite{guo2023simlpe}, and stochastic approaches~\cite{barquero2023belfusion,chen2023humanmac,sohn2015cvae,ho2020ddpm}.
Recent tokenized and latent-predictive formulations further broaden the modeling space~\cite{jiang2023motiongpt,seff2023motionlm,chen2024motionllm,assran2023ijepa}.
However, standard benchmarks are built primarily from laboratory motion capture~\cite{ionescu2014human36m,mahmood2019amass} or outdoor humans~\cite{vonmarcard20183dpw,adeli2021tripod,waymo3dskelmo2025,vcpose2026}, rather than seated, partially occluded drivers observed from heterogeneous cabin viewpoints.
DriveMotion brings these forecasting paradigms into the in-cabin domain under a common protocol; additional model families are discussed in Appendix~D.

\section{The DriveMotion Dataset}
\label{sec:dataset}

\begin{figure*}[htbp]
  \centering
  \includegraphics[width=0.98\linewidth]{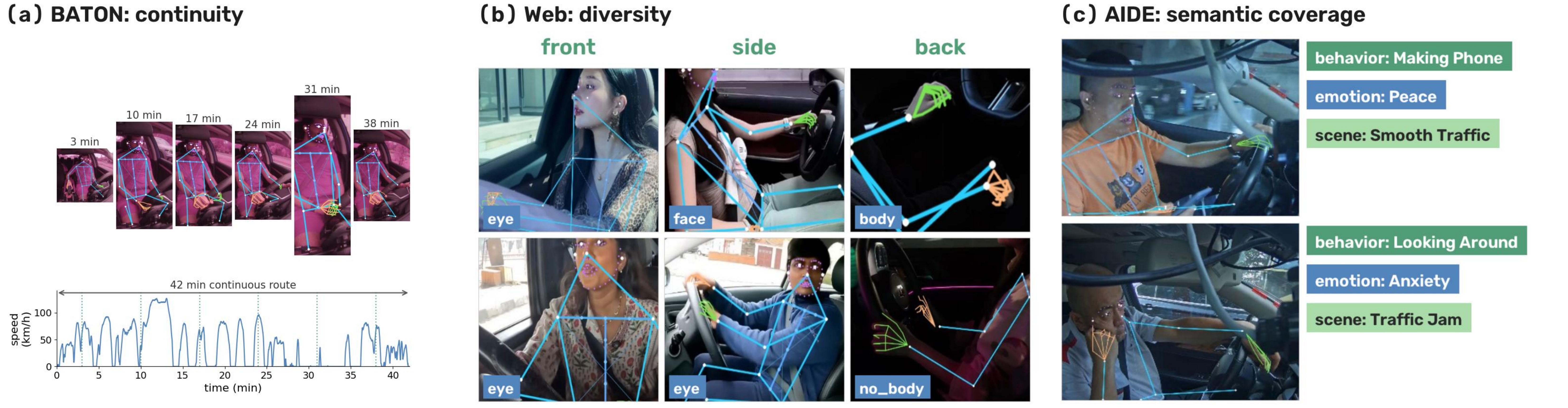}
  \caption{Examples from the three DriveMotion sources.}
  \label{fig:examples}
\end{figure*}

The forecasting benchmark in \cref{sec:benchmark} requires temporal continuity, observation variation, and semantic labels for activity-based analysis. To cover these requirements more comprehensively, DriveMotion combines three complementary sources (\cref{fig:examples}).

\subsection{Data sources}
\label{sec:sources}

\paragraph{BATON: continuity.}
BATON~\cite{baton2026} is a naturalistic corpus recorded by consumer driving devices~\cite{schafer2018comma2k19} from \DMbatonDrivers{} drivers. 
Each route provides a continuous cabin-facing fisheye stream, a road-facing stream, on-device head pose, and time-aligned CAN.
BATON includes \DMbatonRoutes{} routes spanning minutes to hours, totaling \DMbatonHours{} hours of continuous driving data (\cref{fig:examples}a).
The road-facing video is released as time-aligned exterior-context features, an optional model input (\cref{sec:task}). 
The CAN channels are used offline to locate maneuver intervals during benchmark construction and for hardware-grounded head-pose evaluation. A fixed-mount fleet cannot supply variation in observation, since every driver is recorded from the same mount.

\paragraph{Web corpus: observation variation.}
Cabin footage in the wild varies in where the camera sits and how much of the driver it sees. We curate publicly available self-recorded driving videos from \DMwebCreators{} creators (\DMwebDrivers{} drivers) and annotate them in-house: \DMwebAnnotClips{} candidate clips receive an eleven-class view label, and each of the \DMwebDriverClips{} driver clips additionally receives a viewpoint label (front/side/back), a five-level visibility label (\emph{eye} $\supset$ \emph{face} $\supset$ \emph{head} $\supset$ \emph{body} $\supset$ \emph{arms-only}), and a reference driver box that anchors extraction. Simultaneously recorded views (road, BEV) are recovered as time-aligned streams and released in the same exterior-context format as the fleet road stream, covering 79\% of web forecasting windows. This corpus supplies the viewpoint and visibility variation that cross-view evaluation requires.

\paragraph{AIDE: semantic coverage.}
We re-map AIDE~\cite{yang2023aide}, \DMaideClips{} multi-view in-cabin clips labeled with driver behavior, emotion, and scene context, into the DriveMotion representation while preserving its annotations (\cref{fig:examples}c). These labels support stratifying forecasting error by driver activity and transfer evaluation of learned motion representations.

\subsection{Unified representation and construction}
\label{sec:pipeline}

Every sequence is a tuple $(X, C, M, h, u, \text{meta})$ on a fixed 10\,Hz grid: keypoints $X = (x_t)$ with $x_t \in \mathbb{R}^{J \times 2}$ in normalized image coordinates over the $J{=}133$-slot COCO-WholeBody layout~\cite{jin2020cocowholebody}, of which 127 joints (17 body, 68 face, 42 hands) are active in-cabin (the six foot joints are never observable from cabin cameras and are permanently masked); confidences $C$; validity masks $M$; head pose $h_t \in \mathbb{R}^3$ (yaw, pitch, roll) with recorded provenance (device-derived on BATON, vision-based elsewhere); and, where available, CAN channels $u_t \in \mathbb{R}^4$ (speed, steering, throttle, brake). Metadata records source, view type, driver, visibility level, and absolute source timestamps. Unobservable joints are represented explicitly through $M$ and are not imputed, so evaluation can condition on partial observability. The primary released visual modality is a rendered skeleton motion video per sequence, which preserves the motion while substantially reducing appearance-based identity information.

All sources pass through one extraction trunk (whole-body pose estimation with RTMW~\cite{jiang2024rtmw,jiang2023rtmpose}, temporal resampling, normalization, validity estimation, and quality scoring) and differ only in how the driver is identified. On BATON, driver association is anchored on the device's reported face position, and a residual detector flags disagreements. On web video, a fixed curation funnel verifies driver presence~\cite{redmon2016yolo}, viewpoint consistency, and cross-frame driver association, applies visibility-aware acceptance, and decomposes composited multi-camera frames into per-view streams. Every rejection is logged with its stage and gate values. AIDE is re-extracted with the same trunk after its shipped pose confidences proved uncalibrated on cabin footage (\cref{sec:quality}). The full pipeline, funnel statistics, and thresholds are in Appendix~B.

\subsection{Overview, quality, and release}
\label{sec:stats}

DriveMotion comprises \DMtotalSeqs{} forecasting sequences (1{,}329 released BATON routes + 3{,}038 web spans) plus \DMaideClips{} AIDE clips, \DMtotalHours{} hours of motion in total, and \DMtotalDrivers{} drivers. BATON contributes \DMbatonHours{}\,h of continuous routes, the web corpus \DMwebHours{}\,h across front, side, and back viewpoints and five visibility levels, and AIDE \DMaideHours{}\,h of annotated clips. Composition statistics and the curation funnel are in Appendix~C. \cref{tab:comparison} compares DriveMotion with existing driver-facing datasets; among those compared, DriveMotion uniquely combines long continuous recordings, whole-body motion sequences, and a forecasting protocol.

\label{sec:quality}
We validate the extracted motion through five complementary audits. On BATON, every extracted skeleton is checked against the device's face-position channel, identifying and removing 12 of 1{,}395 raw routes (0.9\%); extraction and quality gates remove a further 54, leaving the 1{,}329 released routes (full accounting in Appendix~A). Vision-based head pose is evaluated against the device reference, providing a hardware-grounded estimate of the error expected for the vision-only head pose used by the other sources. We further compare annotated visibility levels with extracted keypoint validity and find that joints expected to be outside the visible region are nearly always absent. Motion integrity is assessed on a 3{,}000-window sample: 98.2\% of frame-to-frame trunk-keypoint transitions are below 0.25 shoulder widths (99.4\% below 0.5), the corresponding rate for arm keypoints, the noisiest channel, is 95.6\%, and core-joint validity averages 95.0\%. Per-sequence motion-energy profiles show predominantly low-amplitude motion punctuated by short excursions, motivating the benchmark design in \cref{sec:benchmark}. This audit also revealed uncalibrated pose confidences from a prior pipeline~\cite{fang2023alphapose} on cabin footage, leading us to re-extract AIDE with the unified pipeline. Finally, we release the complete web curation funnel, including per-stage rejection causes and gate values, so that the filtering process can be audited.

\label{sec:scope}
\textbf{Release and privacy.} For BATON we release motion sequences with head pose and permitted CAN channels. For the web corpus we release derived skeleton sequences, skeleton motion videos, and the curation ledger, with no identity-preserving crops of the driver. For AIDE we release derived annotations and extraction scripts that reproduce our sequences from an official copy. Skeleton-derived artifacts are CC~BY~4.0; aligned context-view clips are released separately under a research-only license. Splits are grouped by recording device (BATON) and creator channel (web) to prevent driver leakage; motion itself can carry identity cues, which is why disjoint evaluation is enforced at this grouping. DriveMotion contains no metric 3D ground truth and no synchronized multi-angle captures of the same driver, and fisheye distortion is left in place and documented. The collection recipe can grow: the fleet keeps recording, the web pipeline can ingest new public videos, and further datasets can be re-mapped as AIDE was.

\section{Forecasting Benchmark}
\label{sec:benchmark}


This section defines the forecasting task and its evaluation settings. Given observed driver motion, models predict future keypoint trajectories and head pose over the next several seconds. We evaluate forecasting in two settings: maneuver-centered windows identified offline from CAN signals, and the natural window distribution used to test generalization across data sources and camera views. We then describe the evaluation metrics and baseline models.

\begin{figure*}[htbp]
  \centering
  \includegraphics[width=0.9\linewidth]{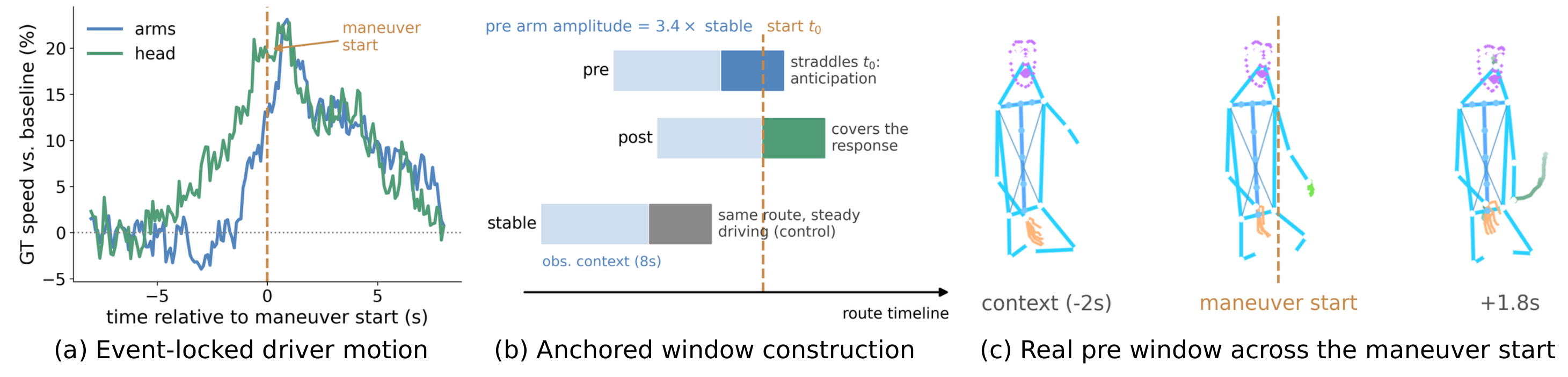}
  \caption{The dynamics-anchored protocol. Green trails in (c) trace wrist/nose motion over the forecast interval.}
  \label{fig:protocol}
\end{figure*}

\subsection{Task and splits}
\label{sec:task}

Given 8\,s of observed driver motion $(x_t, c_t, m_t, h_t)_{t=1}^{80}$, a model predicts the next 4\,s of keypoint trajectories and head pose at 10\,Hz in a canonical torso frame. Models may optionally use synchronized exterior-scene features $S$ over the observed interval. These features are 2\,Hz frozen ResNet-50 embeddings from the BATON road-facing camera and the recovered front and bird's-eye views in the web corpus, covering 96\% and 79\% of windows, respectively. Missing features are set to zero with an availability flag. The exterior stream is limited to the observation window and never includes future frames.
Every architecture is also evaluated using pose alone to measure how much future motion can be predicted from motion history. CAN signals are used only offline to identify maneuver intervals during benchmark construction and are never provided to models at inference. Benchmark windows are sampled at a fixed stride, yielding \DMtotalWindows{} windows. Because AIDE clips are shorter than the full observation-and-forecast interval, forecasting uses BATON and the web corpus, while AIDE is retained for semantic transfer evaluation. Splits are identity-disjoint, grouped by recording device for BATON and creator channel for the web corpus, and fixed before event mining so that no driver appears across split boundaries.


\subsection{Protocol A: dynamics-anchored forecasting}
\label{sec:protoA}

Under uniform sampling, most windows contain little driver motion, making persistence a strong baseline (Sec.~\ref{sec:results}). Protocol A instead evaluates forecasting around vehicle maneuvers. Turns, lane changes, braking, acceleration, and stops are identified offline from CAN signals, yielding 76{,}026 maneuver initiations across all 1{,}395 raw BATON routes (CAN is usable on every raw route, independent of the quality gates of \cref{sec:quality}). These maneuvers are used only to select evaluation windows; future driver motion is never used in window selection. Driver motion increases around these transitions (Fig.~\ref{fig:protocol}a): head motion begins to rise before maneuver onset, arm motion peaks near execution, and arm amplitude in pre-maneuver windows is $3.4\times$ that in matched stable-driving controls.

Each maneuver start $t_0$ defines a pre-window and a post-window. The pre-window forecast interval spans $t_0{\pm}2$\,s, covering the transition from pre-maneuver motion into execution, while the post-window forecast covers $[t_0, t_0{+}4\,\text{s}]$. Stable controls are sampled from the same routes during steady driving, and windows with an essentially stationary vehicle are excluded from all three groups. The anchored test set contains 6{,}138 pre, 6{,}238 post, and 4{,}262 stable windows. Results are reported separately for each group and as a macro average. During training, batches are maneuver-enriched by mixing anchored windows with windows from the natural distribution in roughly equal proportions. Anchored training windows are constructed from training routes using the same procedure as the test windows. The same training setup is applied to all baselines, and the effect of enrichment is examined through a dose--response study (Sec.~\ref{sec:results}).

\subsection{Protocol B: robustness under observation shift}
\label{sec:protoB}
\label{sec:protocols}

Protocol B measures whether a model trained in one acquisition environment generalizes to differently recorded driver imagery. It retains the natural window distribution, since generalization should be measured on the distribution deployed models will encounter. The web corpus serves as the held-out evaluation domain: the same architecture trained on the fleet source alone is compared against training on the full corpus, evaluated per source, with breakdowns by annotated viewpoint (front/side/back) and five-level visibility (Appendix~L).

\subsection{Metrics}
\label{sec:metrics}

We use two primary metrics. \textbf{MPJPE@4s} measures geometric accuracy as the mean per joint error at 4\,s. It is computed over 23 evaluation points, including the nose, eyes, ears, shoulders, hips, and jaw line, which are observable across all camera placements. Errors are computed in the canonical torso frame using the validity mask $M$ and are reported in units of $10^{-2}$ shoulder widths.

\textbf{Part-State F1@2s} measures future part-level motion state. Predicted trajectories are converted into three motion states, \emph{still}, \emph{small}, and \emph{large}, for the head, torso, and each arm. State thresholds are fixed using training-split percentiles, so the three categories are dataset-relative rather than absolute motion classes, and predictions are scored with macro F1 against the ground truth states. The two metrics capture different aspects of forecasting. MPJPE uses a 23 point set that supports comparison across camera views, while Part-State F1 also evaluates arm motion, which cannot be included reliably in a common point set across all views.

Head pose is evaluated only where hardware reference labels are available. Vision based head pose on web footage produces spurious excursions in 97\% of windows, compared with a 6.1\% rate for the hardware reference, so evaluating against the vision based estimates would largely measure annotation noise.

We also report freeze rate to identify forecasts that collapse toward stillness. Freeze rate is the fraction of predictions whose net displacement falls below a threshold calibrated from the ground truth. We use net displacement rather than frame level displacement because frame to frame motion is strongly affected by keypoint jitter. With this definition, the ground truth frozen fraction is 0.057, which provides a reference for interpreting model freeze rates.

Additional forecasting horizons, region errors, stochastic best of $N$ results, threshold definitions, and statistical procedures are provided in Appendices~E--F. These procedures include fixed hashed subsets, bootstrap confidence intervals clustered by route and event, and the rule that overlapping windows are not treated as independent samples.

\subsection{Baselines}
\label{sec:baselines}

We evaluate persistence (zero-motion) and visibility-guarded constant-velocity baselines together with representative recurrent (GRU~\cite{cho2014gru,martinez2017human}), MLP (siMLPe~\cite{guo2023simlpe}), Transformer~\cite{vaswani2017attention}, generative (CVAE~\cite{sohn2015cvae}, diffusion~\cite{ho2020ddpm,chen2023humanmac}), and motion-token~\cite{jiang2023motiongpt,seff2023motionlm} forecasters under one training harness, each with and without exterior context. 
The motion-token models quantize core-joint displacements into 256 bins per channel and decode autoregressively. 
The Transformer family is additionally scaled from 7.6M to 133M parameters to test capacity effects. 
All models use residual decoding from the last observed pose~\cite{martinez2017human}, which we found necessary to avoid mean-collapse. 
Architecture and optimization details, JEPA-, LLM-, and world-model-style diagnostic baselines, and objective-level baselines (calibrated decoding, RL fine-tuning) are in Appendix~G; probe analyses (maneuver-intent anticipation, semantic transfer) are in Appendix~N.

\section{Results}
\providecommand{\std}[1]{{\scriptsize$\pm$#1}}
\label{sec:results}

We report anchored forecasting (Sec.~\ref{sec:res_learnable}), generalization under observation shift (Sec.~\ref{sec:crosssource}), and the relation between geometric and behavioral accuracy (Sec.~\ref{sec:res_headroom}). Extended results are in Appendices~H--P.

\begin{table*}[t]
\centering
\caption{
Dynamics-anchored forecasting (Protocol A);
macro = stratum mean,
\textsuperscript{p} = pose-only,
$\dagger$ = adapted implementation.
Bold denotes the best macro result on each primary metric.
Additional variants and diagnostics are in Appendix~I.
}
\label{tab:dynmain}
\small
\setlength{\tabcolsep}{4.0pt}
\adjustbox{max width=\textwidth}{%
\rowcolors{3}{white}{dmblue3}
\begin{tabular}{@{\extracolsep{\fill}}lcccccccc}
\toprule
 & \multicolumn{4}{c}{MPJPE@4s $\downarrow$}
 & \multicolumn{4}{c}{Part-State F1@2s $\uparrow$} \\
Model & Pre & Post & Stable & Macro & Pre & Post & Stable & Macro \\
\midrule

Zero-motion
& 8.85 & 9.07 & 5.33 & 7.75
& 0.196 & 0.180 & 0.269 & 0.215 \\

\midrule

GRU seq2seq
& 7.84{\scriptsize$\pm$0.01}
& 8.00{\scriptsize$\pm$0.01}
& 4.67{\scriptsize$\pm$0.01}
& 6.83{\scriptsize$\pm$0.01}
& 0.259{\scriptsize$\pm$0.003}
& 0.245{\scriptsize$\pm$0.002}
& 0.322{\scriptsize$\pm$0.001}
& 0.275{\scriptsize$\pm$0.001} \\

siMLPe\textsuperscript{p}~\cite{guo2023simlpe}
& 7.82{\scriptsize$\pm$0.01}
& 8.07{\scriptsize$\pm$0.00}
& 4.66{\scriptsize$\pm$0.00}
& 6.85{\scriptsize$\pm$0.00}
& 0.209{\scriptsize$\pm$0.001}
& 0.194{\scriptsize$\pm$0.001}
& 0.278{\scriptsize$\pm$0.001}
& 0.227{\scriptsize$\pm$0.001} \\

Transformer ED
& 7.76{\scriptsize$\pm$0.01}
& 7.94{\scriptsize$\pm$0.02}
& 4.56{\scriptsize$\pm$0.01}
& 6.75{\scriptsize$\pm$0.01}
& 0.265{\scriptsize$\pm$0.001}
& 0.252{\scriptsize$\pm$0.003}
& 0.328{\scriptsize$\pm$0.002}
& 0.282{\scriptsize$\pm$0.002} \\

Transformer ED (enriched)
& 7.99{\scriptsize$\pm$0.05}
& 8.18{\scriptsize$\pm$0.04}
& 4.68{\scriptsize$\pm$0.01}
& 6.95{\scriptsize$\pm$0.03}
& 0.299{\scriptsize$\pm$0.002}
& 0.289{\scriptsize$\pm$0.001}
& 0.338{\scriptsize$\pm$0.003}
& 0.309{\scriptsize$\pm$0.001} \\

Transformer-L\textsuperscript{p}
& 7.63{\scriptsize$\pm$0.00}
& 7.82{\scriptsize$\pm$0.00}
& 4.44{\scriptsize$\pm$0.01}
& 6.63{\scriptsize$\pm$0.00}
& 0.269{\scriptsize$\pm$0.000}
& 0.256{\scriptsize$\pm$0.001}
& 0.335{\scriptsize$\pm$0.001}
& 0.287{\scriptsize$\pm$0.001} \\

Transformer-XL\textsuperscript{p}
& 7.64 & 7.81 & 4.42 & \textbf{6.62}
& 0.266 & 0.258 & 0.329 & 0.284 \\

\midrule

CVAE
& 7.85{\scriptsize$\pm$0.01}
& 8.01{\scriptsize$\pm$0.01}
& 4.65{\scriptsize$\pm$0.01}
& 6.84{\scriptsize$\pm$0.01}
& 0.240{\scriptsize$\pm$0.002}
& 0.227{\scriptsize$\pm$0.002}
& 0.303{\scriptsize$\pm$0.001}
& 0.257{\scriptsize$\pm$0.002} \\

DDPM (enriched)
& 9.83{\scriptsize$\pm$0.00}
& 10.00{\scriptsize$\pm$0.01}
& 6.75{\scriptsize$\pm$0.01}
& 8.86{\scriptsize$\pm$0.00}
& 0.292{\scriptsize$\pm$0.002}
& 0.283{\scriptsize$\pm$0.003}
& 0.323{\scriptsize$\pm$0.001}
& 0.299{\scriptsize$\pm$0.001} \\

Motion-token AR-LM\textsuperscript{p} (enriched)
& 9.28 & 9.52 & 5.68 & 8.16
& 0.458 & 0.454 & 0.461 & \textbf{0.458} \\

LLM (Llama-3B)\textsuperscript{p}
& 9.34 & 9.63 & 5.74 & 8.24
& 0.456 & 0.451 & 0.464 & 0.457 \\

\midrule

JEPA\textsuperscript{p}~\cite{assran2023ijepa}
& 8.62 & 8.74 & 5.27 & 7.54
& 0.235 & 0.220 & 0.296 & 0.250 \\

Driver-WM\textsuperscript{p}~\cite{driverwm2026}
& 8.86 & 9.08 & 5.34 & 7.76
& 0.196 & 0.180 & 0.269 & 0.215 \\

\bottomrule
\end{tabular}}
\end{table*}

\subsection{Anchored forecasting}
\label{sec:res_learnable}
\label{sec:whyanchor}
\label{sec:mainresults}

Under uniform sampling, persistence reaches 10.2 MPJPE@4s and the best learned model 9.0, a 12\% margin; extrapolating instantaneous velocity amplifies keypoint jitter into a $7\times$ larger error. The learned margin is concentrated in the windows that contain motion. On the motion-active quartile of the same test set, the advantage over persistence doubles (19.91${\to}$17.45, $-2.46$ versus $-1.20$ overall), and on the quiet half every model ties with persistence (Table~\ref{tab:active}). The quiet half and active quartile are defined by ground-truth displacement quantiles on the natural test distribution, so the stratification itself involves no model output (full natural-distribution board in Appendix~H).

\begin{table}[htbp]
\centering
\caption{MPJPE@4s on quiet vs.\ motion-active windows.}
\label{tab:active}
\small
\setlength{\tabcolsep}{6pt}
\adjustbox{max width=\linewidth}{%
\rowcolors{2}{white}{dmblue3}
\begin{tabular}{@{\extracolsep{\fill}}lcc}
\toprule
Model & Quiet half & Active quartile \\
\midrule
Zero-motion & 4.89 & 19.91 \\
GRU seq2seq & 5.13 & 18.01 \\
siMLPe & 5.14 & 17.62 \\
Transformer ED & \textbf{4.79} & \textbf{17.45} \\
CVAE & 5.06 & 17.86 \\
\bottomrule
\end{tabular}}
\end{table}

Protocol A evaluates on these intervals directly. Its premise is verified on ground truth before any model is scored: arm amplitude in pre-maneuver windows is $3.4\times$ that in stable controls drawn from the same routes, which holds driver, vehicle, and route composition fixed, and head motion rises seconds before the maneuver start, consistent with anticipatory visual checking (Fig.~\ref{fig:protocol}a). Window selection uses vehicle state only and never inspects the driver-motion target, and splits are fixed before event mining, so anchoring cannot leak split information.

In Table~\ref{tab:dynmain}, learned models improve over persistence by 13--15\% (macro 7.75${\to}$6.75 for the base Transformer, 6.63 and 6.62 for L and XL). The groups order consistently for every model: pre and post are harder than stable in absolute error, and the pre stratum places the forecast onset before the maneuver, so the model must bridge from pre-maneuver motion into execution. Evidence that pre-action anticipatory signal exists at all comes from the intent probe (Sec.~\ref{sec:res_headroom}). Part-State F1 improves from 0.215 to 0.309 over the same models. Forecasting skill on this data is concentrated in motion-active, maneuver-related intervals, and evaluation needs to resolve this structure.

Enrichment concentrates its gains in the pre and post groups, raising Part-State F1 from 0.265 to 0.299 and from 0.252 to 0.289, respectively, at a 0.20 macro MPJPE cost. Generative and token-based models reveal a complementary trade-off: DDPM improves motion-state prediction but sacrifices coordinate accuracy, whereas the motion-token AR-LM achieves the highest raw Part-State F1 (0.458) despite substantially higher MPJPE. Additional variants are reported in Appendix~I.

\subsection{Generalization under observation shift}
\label{sec:crosssource}

Protocol B keeps the natural window distribution and evaluates on held-out web creators; anchored evaluation is fleet-only by construction, since the anchors require CAN. A Transformer trained only on the fleet source performs well in-domain (6.04 @4s) but reaches 19.68 on held-out web windows, 45\% worse than repeating the last pose (Table~\ref{tab:crosssource}). Training on the full corpus reduces web error to 12.18, a 38\% reduction and the only configuration that beats persistence there, while remaining close to fleet-only in-domain performance (6.13 vs. 6.04). The held-out set consists of unseen creators within the web source, so this measures robustness to observation shift, not zero-shot transfer to an unseen source. The effect is largest on front views, the viewpoint farthest from the fleet camera (24.0${\to}$15.4). Well-observed web groups behave like fleet data, and heavily occluded groups are dominated by partial observability (per-view and per-visibility breakdowns in Appendix~L). Head-pose results depend on label quality: on the fleet subset with production DMS head pose, learned models beat head persistence (4.19$^\circ$ vs.\ 4.80$^\circ$@4s), while on vision-derived web labels no model does (Appendix~L). The zero-motion reference itself rises from 7.14 on fleet windows to 13.59 on web windows, so part of the web difficulty is intrinsic to the footage; the fleet-only model exceeds even this reference, which indicates learned features tied to the fleet viewpoint. The full-corpus model uses the same architecture, schedule, and parameter budget as the fleet-only one, so the entire reduction is attributable to training-source composition. For the model families tested here, training-source variation affects cross-view transfer substantially more than architectural scaling.
\begin{table}[htbp]
\small
\centering
\caption{Cross-source transfer (MPJPE@4s).}
\label{tab:crosssource}
\small
\setlength{\tabcolsep}{6pt}
\adjustbox{max width=\linewidth}{%
\rowcolors{2}{white}{dmblue3}
\begin{tabular}{@{\extracolsep{\fill}}lcc}
\toprule
Training data & BATON test & Web test \\
\midrule
Zero-motion (reference) & 7.14 & 13.59 \\
BATON only & \textbf{6.04} & 19.68 \\
BATON + Web (full corpus) & 6.13 & \textbf{12.18} \\
\bottomrule
\end{tabular}}
\end{table}

\subsection{Geometric and behavioral accuracy}
\label{sec:res_headroom}
\label{sec:stateresults}
\label{sec:sceneresults}
\label{sec:enriched}


Geometric accuracy and behavioral anticipation favor different models. Transformer-XL obtains the lowest macro MPJPE (6.62), with Transformer-L essentially equal (6.63) at a third of the parameters, while the enriched Transformer attains the best forecast-derived Part-State F1 among regression models (0.309, 44\% above the zero-motion reference) despite higher coordinate error (6.95). Motion-token models further improve state prediction while remaining weaker in coordinates (Table~\ref{tab:dynmain}; full board in Appendix~I). For driver monitoring, this behavioral axis captures whether the head, hand, or torso is about to move.

Three controlled comparisons sit behind the board. Raising the fraction of anchored windows in training improves Part-State F1 monotonically (0.275${\to}$0.291${\to}$0.309); the official 48\% mixture preserves coordinate accuracy, and the 90\% dose trades 4\% of coordinates for further state gains (dose--response in Appendix~I). Pairing each architecture with its pose-only counterpart shows that the frozen exterior features do not improve coordinate accuracy, but improve Part-State F1 under maneuver-enriched training (0.292${\to}$0.309) and reduce freeze rate from 0.51 to 0.40 (full ablation in Appendix~I); stronger scene encoders remain untested. Scaling from 7.6M to 44.5M parameters improves every stratum at negligible seed variance (pose-only pre 7.73${\to}$7.63, stable 4.50${\to}$4.44), and 133M holds these values without extending them (7.64 pre, 4.42 stable; macro 6.63${\to}$6.62), so above roughly 45M the limiting factor on this corpus is no longer capacity. The learned advantage is smaller in the action region: on the natural test set, arm MPJPE improves only from 28.1 (persistence) to 26.8, compared with 10.2 to 9.0 on the stable 23-point set (Appendix~L), which is the benchmark scores arm behavior through Part-State F1 rather than coordinates.

\paragraph{Decoding under-expresses predictive signal.}
Much of the predictable state signal is persistent: a prior that simply carries the observed per-part amplitude forward reaches 0.59--0.65 per-part F1 and outperforms every trajectory-derived predictor (Appendix~K), so raw Part-State F1 partly measures how well a model expresses persisting motion, and the gap between learned predictive information and decoded trajectories is substantial. 
For Transformer-L, a monotone per-part amplitude recalibration fitted only on anchored training windows raises macro Part-State F1 from 0.287 to 0.517 while leaving coordinate metrics unchanged; the trajectory coordinates are not modified, and the calibration applies only to the per-part amplitude used for state decoding. 
This gain is supported by representation-level evidence: a linear probe on frozen forecaster features reaches 0.52--0.58 per-part F1, indicating that useful future motion-state information is already present before trajectory decoding. 
Motion-token models provide complementary evidence, reaching 0.458 raw macro F1 despite substantially worse coordinate accuracy, with similar state performance across different conditioning and backbone variants. 
Together with their strong ordinal ranking performance, these results suggest that current forecasters often identify upcoming motion but under-estimate its amplitude when decoding continuous trajectories. 
Full calibration, per-part, and token analyses are provided in Appendices~J--K; qualitative examples in Appendix~O show the same failure as under-committed arm excursions.

\paragraph{Intent probe.} We label each fleet test window with whether a maneuver of kind $k$ begins within 1--2\,s after the observation ends. A linear probe on 8\,s of observed pose predicts left turns and hard braking at $2.8\times$ and $2.9\times$ AUPRC over prevalence (bootstrap CIs excluding chance) and deceleration and acceleration at $1.5$--$1.9\times$; lane-change precursors are not separable from chance from pose alone (per-kind table in Appendix~N). A classifier trained from scratch with matched capacity performs comparably to the probe, so the anticipatory signal resides in the pose dynamics, not in forecaster-specific features. Pre-maneuver pose therefore carries anticipatory information about some maneuver types and not others, which bounds what any pose-only forecaster can achieve on the remainder.


\begin{figure}[htbp]
  \centering
  \includegraphics[width=0.99\linewidth]{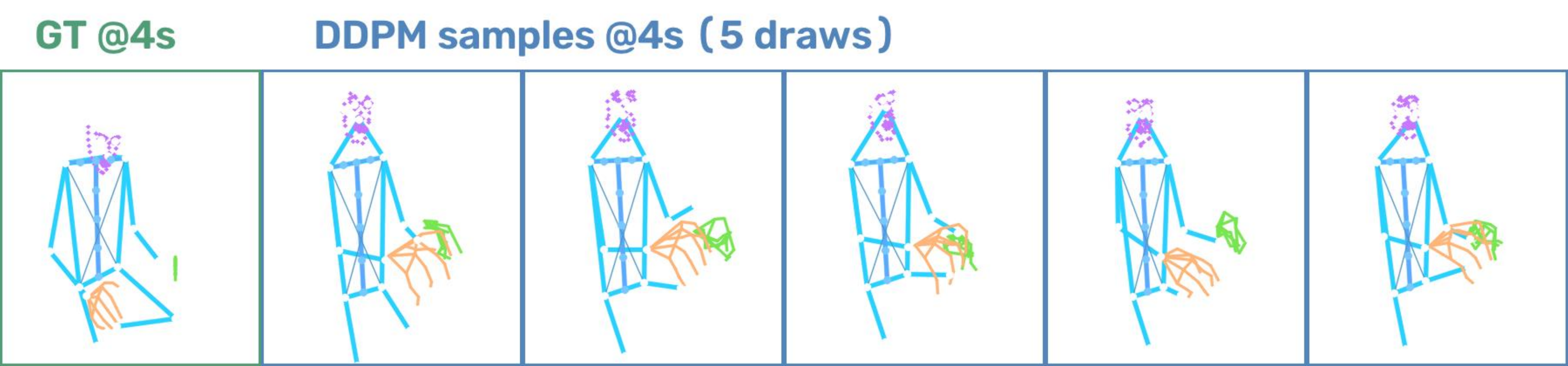}
  \caption{Five DDPM samples at 4\,s against ground truth.}
  \label{fig:diversity}
\end{figure}

\paragraph{Limitations.} DriveMotion evaluates forecasting of a standardized extracted motion representation (RTMW keypoints with validity masks) rather than manually annotated pose ground truth; the audits of Sec.~\ref{sec:quality} check temporal plausibility and cross-sensor consistency, not absolute keypoint accuracy. It contains no metric 3D ground truth and no synchronized multi-angle captures of the same driver, and head-pose provenance differs by source, which is why head metrics are scored on hardware-grounded labels. The web corpus reflects creator-selected recording practices, so its viewpoint mixture is not a controlled sample of deployment cameras. The released exterior features are deliberately simple, and some behaviors, lane-change precursors in particular, are not predictable from skeleton alone. A reward-time route, policy-gradient fine-tuning of the token model with the Part-State metric as reward, did not improve over its supervised baseline and is reported as an open direction (Appendix~G).

\section{Conclusion}
\label{sec:conclusion}

Driver motion is dominated by persistence in ordinary driving and becomes markedly more predictable around maneuvers, where learned models gain 13--15\% over persistence; 
robustness under observation shift depends strongly on training-source variation, with 38\% lower error on held-out web drivers.
Current models capture part of this signal but under-express behaviorally salient motion, and closing the gap between ranked and decoded motion states is the main open modeling problem the benchmark measures. DriveMotion provides the corpus, protocols, and reference implementations needed to study when driver motion is predictable and how that predictability should be evaluated. The release includes identity-disjoint splits, the fixed evaluation subsets, frozen state thresholds, the curation ledger, and training configurations for every baseline, so reported numbers can be reproduced and extended without re-deriving the protocol.

\clearpage
{\small
\bibliographystyle{ieeenat_fullname}
\bibliography{references}

@article{baton2026,
  author={Anonymous},
  title={{BATON}: A Multimodal Benchmark for Bidirectional Automation Transition Observation in Naturalistic Driving},
  journal={arXiv preprint arXiv:2604.07263},
  year={2026}
}

@inproceedings{yang2023aide,
  author={Yang, Dingkang and Huang, Shuai and Xu, Zhi and Li, Zhenpeng and Wang, Shunli and Li, Mingcheng and Wang, Yuzheng and Liu, Yang and Yang, Kun and Chen, Zhaoyu and Wang, Yan and Liu, Jing and Zhang, Peixuan and Zhai, Peng and Zhang, Lihua},
  title={{AIDE}: A Vision-Driven Multi-View, Multi-Modal, Multi-Tasking Dataset for Assistive Driving Perception},
  booktitle={ICCV},
  year={2023}
}

@inproceedings{martin2019driveandact,
  author={Martin, Manuel and Roitberg, Alina and Haurilet, Monica and Horne, Matthias and Rei{\ss}, Simon and Voit, Michael and Stiefelhagen, Rainer},
  title={Drive\&Act: A Multi-Modal Dataset for Fine-Grained Driver Behavior Recognition in Autonomous Vehicles},
  booktitle={ICCV},
  year={2019}
}

@inproceedings{ortega2020dmd,
  author={Ortega, Juan Diego and Kose, Neslihan and Ca{\~n}as, Paola and Chao, Min-An and Unnervik, Alexander and Nieto, Marcos and Otaegui, Oihana and Salgado, Luis},
  title={{DMD}: A Large-Scale Multi-Modal Driver Monitoring Dataset for Attention and Alertness Analysis},
  booktitle={ECCV Workshops},
  year={2020}
}

@article{jain2021mdmd,
  title={The multimodal driver monitoring database: A naturalistic corpus to study driver attention},
  author={Jha, Sumit and Marzban, Mohamed F and Hu, Tiancheng and Mahmoud, Mohamed H and Al-Dhahir, Naofal and Busso, Carlos},
  journal={IEEE Transactions on Intelligent Transportation Systems},
  volume={23},
  number={8},
  pages={10736--10752},
  year={2021},
  publisher={IEEE}
}

@inproceedings{ramanishka2018hdd,
  author={Ramanishka, Vasili and Chen, Yi-Ting and Misu, Teruhisa and Saenko, Kate},
  title={Toward Driving Scene Understanding: A Dataset for Learning Driver Behavior and Causal Reasoning},
  booktitle={CVPR},
  year={2018}
}

@article{wang2023hundreddriver,
  author={Wang, Jing and Li, Wenjing and Li, Fang and Zhang, Jun and Wu, Zhongcheng and Zhong, Zhun and Sebe, Nicu},
  title={100-Driver: A Large-Scale, Diverse Dataset for Distracted Driver Classification},
  journal={IEEE Transactions on Intelligent Transportation Systems},
  year={2023}
}

@inproceedings{kopuklu2021dad,
  author={K{\"o}p{\"u}kl{\"u}, Okan and Zheng, Jiapeng and Xu, Hang and Rigoll, Gerhard},
  title={Driver Anomaly Detection: A Dataset and Contrastive Learning Approach},
  booktitle={WACV},
  year={2021}
}

@article{daos2026,
  author={Anonymous},
  title={{DAOS}: A Multimodal In-Cabin Behavior Monitoring with Driver Action-Object Synergy Dataset},
  journal={IEEE Transactions on Intelligent Transportation Systems},
  year={2026}
}

@article{nobari2024mand,
  author={Dargahi Nobari, Khazar and Bertram, Torsten},
  title={A Multimodal Driver Monitoring Benchmark Dataset for Driver Modeling in Assisted Driving Automation},
  journal={Scientific Data},
  volume={11},
  year={2024}
}

@inproceedings{roth2019ddpose,
  author={Roth, Markus and Gavrila, Dariu M.},
  title={{DD-Pose} -- A Large-Scale Driver Head Pose Benchmark},
  booktitle={IEEE Intelligent Vehicles Symposium (IV)},
  year={2019}
}

@inproceedings{avdm2024,
  title={Automated vehicle driver monitoring dataset from real-world scenarios},
  author={Sabry, Moharned and Morales-Alvarez, Walter and Olaverri-Monreal, Cristina},
  booktitle={2024 IEEE 27th International Conference on Intelligent Transportation Systems (ITSC)},
  pages={1545--1550},
  year={2024},
  organization={IEEE}
}

@article{driverwm2026,
  title={Driver-WM: A Driver-Centric Traffic-Conditioned Latent World Model for In-Cabin Dynamics Rollout},
  author={Chi, Haozhuang and Qiu, Daosheng and Su, Hao and Liu, Haochen and Li, Zirui and Zhang, Haoruo and Lv, Chen},
  journal={arXiv preprint arXiv:2605.05092},
  year={2026}
}

@inproceedings{jain2016brain4cars,
  author={Jain, Ashesh and Koppula, Hema S. and Soh, Shane and Raghavan, Bharad and Singh, Avi and Saxena, Ashutosh},
  title={{Brain4Cars}: Car That Knows Before You Do via Sensory-Fusion Deep Learning Architecture},
  booktitle={ICCV},
  year={2015},
  note={arXiv:1601.00740}
}

@article{poguise2024,
  title={PO-GUISE+: Pose and object guided transformer token selection for efficient driver action recognition},
  author={Pizarro, Ricardo and Valle, Roberto and Barea, Rafael and Buenaposada, Jos{\'e} M and Baumela, Luis and Bergasa, Luis M},
  journal={IEEE Transactions on Intelligent Transportation Systems},
  year={2026},
  publisher={IEEE}
}

@article{posevinet2023,
  title={PoseViNet: Distracted driver action recognition framework using multi-view pose estimation and vision transformer},
  author={Sengar, Neha and Kumari, Indra and Lee, Jihui and Har, Dongsoo},
  journal={arXiv preprint arXiv:2312.14577},
  year={2023}
}

@inproceedings{ma2023cemformer,
  author={Ma, Yunsheng and Ye, Wenqian and Cao, Xu and Abdelraouf, Amr and Han, Kyungtae and Gupta, Rohit and Wang, Ziran},
  title={{CEMFormer}: Learning to Predict Driver Intentions from In-Cabin and External Cameras via Spatial-Temporal Transformers},
  booktitle={IEEE ITSC},
  year={2023},
  note={arXiv:2305.07840}
}

@article{riskaware2026,
  title={Risk-Aware Selective Multimodal Driver Monitoring with Driver-State World Modeling},
  author={Qiu, Daosheng and Chi, Haozhuang and Su, Hao and Long, Shu and Miao, Xinyue and Dong, Yongle and Zhang, Wei},
  journal={arXiv preprint arXiv:2606.26922},
  year={2026}
}

@article{sdatr2025,
  title={Spatiotemporal decoupling attention transformer for 3D skeleton-based driver action recognition},
  author={Xu, Zhuoyan and Xu, Jingke},
  journal={Complex \& Intelligent Systems},
  volume={11},
  number={4},
  pages={195},
  year={2025},
  publisher={Springer}
}

@inproceedings{peng2022transdarc,
  author={Peng, Kunyu and Roitberg, Alina and Yang, Kailun and Zhang, Jiaming and Stiefelhagen, Rainer},
  title={{TransDARC}: Transformer-Based Driver Activity Recognition with Latent Space Feature Calibration},
  booktitle={IEEE/RSJ IROS},
  year={2022}
}

@inproceedings{martinez2017human,
  author={Martinez, Julieta and Black, Michael J. and Romero, Javier},
  title={On Human Motion Prediction Using Recurrent Neural Networks},
  booktitle={CVPR},
  year={2017}
}

@inproceedings{yan2018stgcn,
  author={Yan, Sijie and Xiong, Yuanjun and Lin, Dahua},
  title={Spatial Temporal Graph Convolutional Networks for Skeleton-Based Action Recognition},
  booktitle={AAAI},
  year={2018}
}

@inproceedings{guo2023simlpe,
  author={Guo, Wen and Du, Yuming and Shen, Xi and Lepetit, Vincent and Alameda-Pineda, Xavier and Moreno-Noguer, Francesc},
  title={Back to {MLP}: A Simple Baseline for Human Motion Prediction},
  booktitle={WACV},
  year={2023}
}

@inproceedings{mao2019learning,
  author={Mao, Wei and Liu, Miaomiao and Salzmann, Mathieu and Li, Hongdong},
  title={Learning Trajectory Dependencies for Human Motion Prediction},
  booktitle={ICCV},
  year={2019}
}

@inproceedings{mao2020history,
  author={Mao, Wei and Liu, Miaomiao and Salzmann, Mathieu},
  title={History Repeats Itself: Human Motion Prediction via Motion Attention},
  booktitle={ECCV},
  year={2020}
}

@inproceedings{chen2023humanmac,
  author={Chen, Ling-Hao and Zhang, Jiawei and Li, Yewen and Pang, Yiren and Xia, Xiaobo and Liu, Tongliang},
  title={{HumanMAC}: Masked Motion Completion for Human Motion Prediction},
  booktitle={ICCV},
  year={2023}
}

@inproceedings{adeli2021tripod,
  author={Adeli, Vida and Ehsanpour, Mahsa and Reid, Ian and Niebles, Juan Carlos and Savarese, Silvio and Adeli, Ehsan and Rezatofighi, Hamid},
  title={{TRiPOD}: Human Trajectory and Pose Dynamics Forecasting in the Wild},
  booktitle={ICCV},
  year={2021}
}

@inproceedings{barquero2023belfusion,
  author={Barquero, German and Escalera, Sergio and Palmero, Cristina},
  title={{BeLFusion}: Latent Diffusion for Behavior-Driven Human Motion Prediction},
  booktitle={ICCV},
  year={2023}
}

@inproceedings{waymo3dskelmo2025,
  title={Waymo-3dskelmo: A multi-agent 3d skeletal motion dataset for pedestrian interaction modeling in autonomous driving},
  author={Zhu, Guangxun and Fan, Shiyu and Dai, Hang and Ho, Edmond SL},
  booktitle={Proceedings of the 33rd ACM International Conference on Multimedia},
  pages={13184--13190},
  year={2025}
}

@article{vcpose2026,
  title={Modeling 3D Pedestrian-Vehicle Interactions for Vehicle-Conditioned Pose Forecasting},
  author={Zhu, Guangxun and Liu, Xuan and Pugeault, Nicolas and Wei, Chongfeng and Ho, Edmond SL},
  journal={arXiv preprint arXiv:2602.08962},
  year={2026}
}

@inproceedings{jiang2023motiongpt,
  author={Jiang, Biao and Chen, Xin and Liu, Wen and Yu, Jingyi and Yu, Gang and Chen, Tao},
  title={{MotionGPT}: Human Motion as a Foreign Language},
  booktitle={NeurIPS},
  year={2023}
}

@inproceedings{seff2023motionlm,
  author={Seff, Ari and Cera, Brian and Chen, Dian and Ng, Mason and Zhou, Aurick and Nayakanti, Nigamaa and Refaat, Khaled S. and Al-Rfou, Rami and Sapp, Benjamin},
  title={{MotionLM}: Multi-Agent Motion Forecasting as Language Modeling},
  booktitle={ICCV},
  year={2023}
}

@inproceedings{abdelfattah2024sjepa,
  author={Abdelfattah, Mohamed and Alahi, Alexandre},
  title={{S-JEPA}: A Joint Embedding Predictive Architecture for Skeletal Action Recognition},
  booktitle={ECCV},
  year={2024}
}

@inproceedings{assran2023ijepa,
  author={Assran, Mahmoud and Duval, Quentin and Misra, Ishan and Bojanowski, Piotr and Vincent, Pascal and Rabbat, Michael and LeCun, Yann and Ballas, Nicolas},
  title={Self-Supervised Learning from Images with a Joint-Embedding Predictive Architecture},
  booktitle={CVPR},
  year={2023}
}

@article{assran2025vjepa2,
  title={V-jepa 2: Self-supervised video models enable understanding, prediction and planning},
  author={Assran, Mido and Bardes, Adrien and Fan, David and Garrido, Quentin and Howes, Russell and Muckley, Matthew and Rizvi, Ammar and Roberts, Claire and Sinha, Koustuv and Zholus, Artem and others},
  journal={arXiv preprint arXiv:2506.09985},
  year={2025}
}

@inproceedings{rabinowitz2018tomnet,
  author={Rabinowitz, Neil C. and Perbet, Frank and Song, H. Francis and Zhang, Chiyuan and Eslami, S. M. Ali and Botvinick, Matthew},
  title={Machine Theory of Mind},
  booktitle={ICML},
  year={2018}
}

@article{fei2026mental,
  author={Fei, Hao and Zhao, Yiran and others},
  title={Mental World Modeling},
  journal={arXiv preprint arXiv:2607.27201},
  year={2026}
}

@inproceedings{chen2024motionllm,
  author={Chen, Ling-Hao and Lu, Shunlin and Zeng, Ailing and Zhang, Hao and Wang, Benyou and Zhang, Ruimao and Zhang, Lei},
  title={{MotionLLM}: Understanding Human Behaviors from Human Motions and Videos},
  booktitle={ICLR},
  year={2025},
  note={arXiv:2405.20340}
}

@article{wang2024motiongpt2,
  title={Motiongpt-2: A general-purpose motion-language model for motion generation and understanding},
  author={Wang, Yuan and Huang, Di and Zhang, Yaqi and Ouyang, Wanli and Jiao, Jile and Feng, Xuetao and Xu, Dan and Tang, Shixiang},
  journal={IEEE Transactions on Circuits and Systems for Video Technology},
  year={2026},
  publisher={IEEE}
}

@article{bardes2024vjepa,
  author={Bardes, Adrien and Garrido, Quentin and Ponce, Jean and Chen, Xinlei and Rabbat, Michael and LeCun, Yann and Mairal, Julien and Ballas, Nicolas},
  title={Revisiting Feature Prediction for Learning Visual Representations from Video},
  journal={Transactions on Machine Learning Research},
  year={2024}
}

@inproceedings{li2025llamo,
  author={Li, Lei and Jia, Sen and Wang, Jianhao and Jiang, Zhongyu and Zhou, Feng and others},
  title={Human Motion Instruction Tuning},
  booktitle={CVPR},
  year={2025},
  note={arXiv:2411.16805}
}

@article{baharani2025mofm,
  author={Baharani, Mohammadreza and others},
  title={{MoFM}: A Large-Scale Human Motion Foundation Model},
  journal={arXiv preprint arXiv:2502.05432},
  year={2025}
}

@article{jiang2023rtmpose,
  author={Jiang, Tao and Lu, Peng and Zhang, Li and Ma, Ningsheng and Han, Rui and Lyu, Chengqi and Li, Yining and Chen, Kai},
  title={{RTMPose}: Real-Time Multi-Person Pose Estimation Based on {MMPose}},
  journal={arXiv preprint arXiv:2303.07399},
  year={2023}
}

@article{jiang2024rtmw,
  author={Jiang, Tao and Xie, Xinchen and Li, Yining},
  title={{RTMW}: Real-Time Multi-Person 2D and 3D Whole-Body Pose Estimation},
  journal={arXiv preprint arXiv:2407.08634},
  year={2024}
}

@inproceedings{jin2020cocowholebody,
  author={Jin, Sheng and Xu, Lumin and Xu, Jin and Wang, Can and Liu, Wentao and Qian, Chen and Ouyang, Wanli and Luo, Ping},
  title={Whole-Body Human Pose Estimation in the Wild},
  booktitle={ECCV},
  year={2020}
}

@inproceedings{redmon2016yolo,
  author={Redmon, Joseph and Divvala, Santosh and Girshick, Ross and Farhadi, Ali},
  title={You Only Look Once: Unified, Real-Time Object Detection},
  booktitle={CVPR},
  year={2016}
}

@article{schafer2018comma2k19,
  author={Schafer, Harald and Santana, Eder and Haden, Andrew and Biasini, Riccardo},
  title={A Commute in Data: The comma2k19 Dataset},
  journal={arXiv preprint arXiv:1812.05752},
  year={2018}
}

@article{ionescu2014human36m,
  author={Ionescu, Catalin and Papava, Dragos and Olaru, Vlad and Sminchisescu, Cristian},
  title={{Human3.6M}: Large Scale Datasets and Predictive Methods for 3D Human Sensing in Natural Environments},
  journal={IEEE TPAMI},
  year={2014}
}

@inproceedings{mahmood2019amass,
  author={Mahmood, Naureen and Ghorbani, Nima and Troje, Nikolaus F. and Pons-Moll, Gerard and Black, Michael J.},
  title={{AMASS}: Archive of Motion Capture as Surface Shapes},
  booktitle={ICCV},
  year={2019}
}

@inproceedings{vonmarcard20183dpw,
  author={von Marcard, Timo and Henschel, Roberto and Black, Michael J. and Rosenhahn, Bodo and Pons-Moll, Gerard},
  title={Recovering Accurate 3D Human Pose in the Wild Using {IMUs} and a Moving Camera},
  booktitle={ECCV},
  year={2018}
}

@inproceedings{vaswani2017attention,
  author={Vaswani, Ashish and Shazeer, Noam and Parmar, Niki and Uszkoreit, Jakob and Jones, Llion and Gomez, Aidan N. and Kaiser, Lukasz and Polosukhin, Illia},
  title={Attention Is All You Need},
  booktitle={NeurIPS},
  year={2017}
}

@inproceedings{cho2014gru,
  author={Cho, Kyunghyun and van Merri{\"e}nboer, Bart and Gulcehre, Caglar and Bahdanau, Dzmitry and Bougares, Fethi and Schwenk, Holger and Bengio, Yoshua},
  title={Learning Phrase Representations Using {RNN} Encoder-Decoder for Statistical Machine Translation},
  booktitle={EMNLP},
  year={2014}
}

@inproceedings{ho2020ddpm,
  author={Ho, Jonathan and Jain, Ajay and Abbeel, Pieter},
  title={Denoising Diffusion Probabilistic Models},
  booktitle={NeurIPS},
  year={2020}
}

@inproceedings{sohn2015cvae,
  author={Sohn, Kihyuk and Lee, Honglak and Yan, Xinchen},
  title={Learning Structured Output Representation Using Deep Conditional Generative Models},
  booktitle={NeurIPS},
  year={2015}
}

@article{fang2023alphapose,
  author={Fang, Hao-Shu and Li, Jiefeng and Tang, Hongyang and Xu, Chao and Zhu, Haoyi and Xiu, Yuliang and Li, Yong-Lu and Lu, Cewu},
  title={{AlphaPose}: Whole-Body Regional Multi-Person Pose Estimation and Tracking in Real-Time},
  journal={IEEE TPAMI},
  year={2023}
}

@inproceedings{hu2022lora,
  title={Lo{RA}: Low-rank adaptation of large language models},
  author={Hu, Edward J and Shen, Yelong and Wallis, Phillip and Allen-Zhu, Zeyuan and Li, Yuanzhi and Wang, Shean and Wang, Lu and Chen, Weizhu},
  booktitle={International Conference on Learning Representations},
  year={2022}
}

@article{chen2025cockpitllama,
  author  = {Yi Chen and Chengzhe Li and Qirui Yuan and Jinyu Li and
             Yuze Fan and Xiaojun Ge and Yun Li and Fei Gao and Rui Zhao},
  title   = {Cockpit-Llama: Driver Intent Prediction in Intelligent Cockpit
             via Large Language Model},
  journal = {Sensors},
  volume  = {25},
  number  = {1},
  pages   = {64},
  year    = {2025},
  doi     = {10.3390/s25010064},
  url     = {https://doi.org/10.3390/s25010064}
}
}

\clearpage
\appendix
\providecommand{\std}[1]{{\scriptsize$\pm$#1}}

\noindent This appendix covers dataset construction and statistics (Secs.~\ref{sec:app_accounting}--\ref{sec:app_extrelated}), benchmark specification (Secs.~\ref{sec:app_protocol}--\ref{sec:app_training}), extended results (Secs.~\ref{sec:app_natboard}--\ref{sec:app_qual}), diagnostics (Sec.~\ref{sec:app_diag}), and reproducibility (Sec.~\ref{sec:app_repro}).

\section{BATON route accounting}
\label{sec:app_accounting}
All BATON counts in the paper trace to one accounting: 1{,}395 raw fleet routes were processed; every raw route carries usable CAN and therefore contributes to the CAN event bank (76{,}026 events over 1{,}395 routes), independent of whether its motion passed quality control. The driver-association audit removed 12 routes whose extracted skeleton disagreed with the device face-position channel, and extraction, quality, and sequence-assembly gates removed a further 54, leaving the \DMbatonRoutes{} released motion sequences, each with synchronized CAN and device-derived head pose.

\begin{table}[t]
\centering
\caption{BATON route accounting.}
\label{tab:accounting}
\small
\adjustbox{max width=\linewidth}{%
\rowcolors{2}{white}{dmblue3}
\begin{tabular}{@{\extracolsep{\fill}}lr}
\toprule
Stage & Routes \\
\midrule
Raw fleet routes processed (all in CAN event bank) & 1{,}395 \\
Removed by driver-association audit & 12 \\
Removed by extraction / quality / assembly gates & 54 \\
Released motion sequences (CAN + device head pose) & 1{,}329 \\
\bottomrule
\end{tabular}}
\end{table}

\section{Web curation pipeline and outcomes}
\label{sec:app_web}

\begin{figure*}[htbp]
  \centering
  \includegraphics[width=0.92\linewidth]{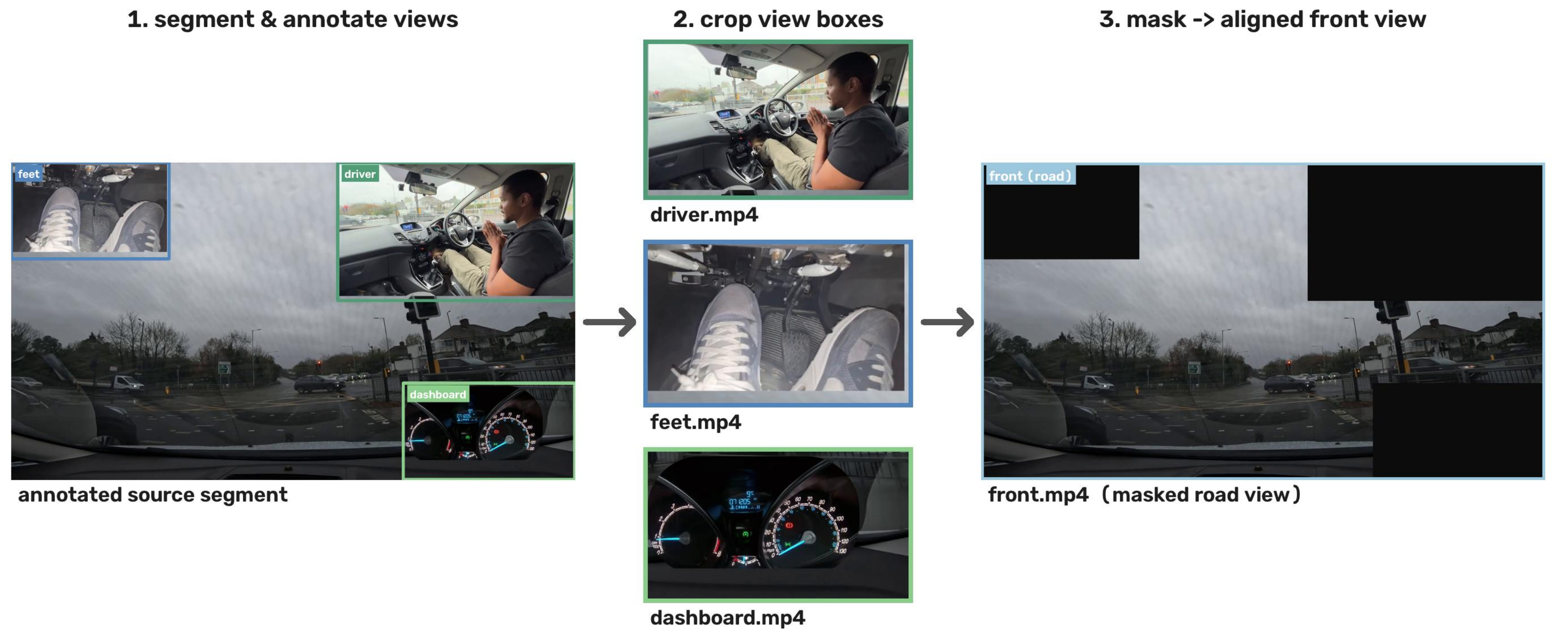}
  \caption{Web curation. Creator-produced driving videos often contain several views composited into one frame. Annotated segments are decomposed into per-view crops (driver, feet, dashboard), and the remaining masked frame yields the aligned road view, so a single source video contributes a synchronized multi-view subset.}
  \label{fig:webpipe}
\end{figure*}

The complete funnel outcome: from 1{,}868 manually annotated clips, 873 driver clips (142.3 source hours) entered extraction; viewpoint verification, driver association, and visibility-aware acceptance retained 4{,}858 continuous spans totalling 79.9 hours, a 56\% retention rate of driver source hours, reflecting deliberately conservative gates; quality gating during sequence assembly further reduces the 4{,}858 accepted spans (79.9\,h) to the 3{,}038 released motion sequences (\DMwebHours{}\,h). Accepted hours by viewpoint: side 60.0, back 13.4, front 6.4; by visibility level: face 36.3, eye 22.9, head 16.9, arms-only 2.3, body 1.6. Every rejection is logged with its stage and gate values in the released curation ledger.

\section{Composition statistics}
\label{sec:app_stats}

\begin{figure*}[htbp]
  \centering
  \includegraphics[width=0.98\linewidth]{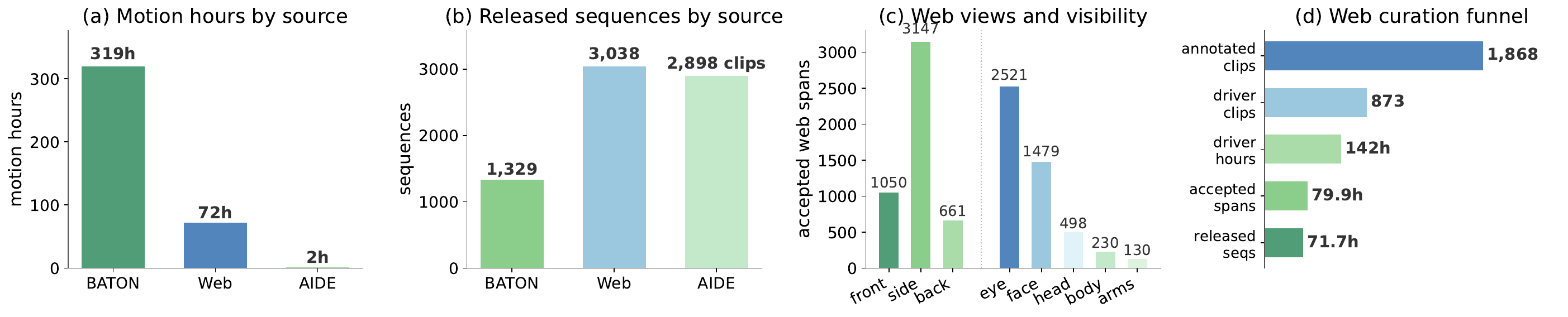}
  \caption{Dataset composition. (a)~Motion hours and (b)~sequence counts show where continuity comes from; (c)~web viewpoint and visibility composition shows that observation diversity is real, not incidental; (d)~the web curation funnel documents how aggressively raw material is filtered before release.}
  \label{fig:stats}
\end{figure*}

The forecasting corpus comprises \DMtotalSeqs{} sequences (1{,}329 BATON + 3{,}038 web), while the semantic transfer subset contributes an additional \DMaideClips{} AIDE clips; together, DriveMotion contains \DMtotalHours{} hours from \DMtotalDrivers{} drivers. \Cref{fig:stats} reads directly against the design goals: continuity is carried by \DMbatonRoutes{} continuous BATON routes (\DMbatonHours{}\,h); observation diversity by the web corpus (\DMwebHours{}\,h across side \DMwebViewSide{}, front \DMwebViewFront{}, and back \DMwebViewBack{} spans and five visibility levels); semantics by \DMaideClips{} AIDE clips (\DMaideHours{}\,h). The curation funnel (\cref{fig:stats}d) is deliberately steep: from \DMwebAnnotClips{} annotated web clips, only driver-verified, viewpoint-stable, quality-gated segments survive. the dataset comparison table in the main text situates DriveMotion among driver-facing datasets: it is the only resource coupling long continuous recordings with released whole-body motion sequences. The conversion of these sequences into standardized observe-and-forecast windows is defined by the benchmark protocol in the main text (Sec.~4).

\section{Extended related work}
\label{sec:app_extrelated}

\paragraph{Motion foundation models and predictive world modeling.}
Recent work increasingly treats human motion as a general sequence-modeling problem rather than a task-specific prediction target.
MotionGPT~\cite{jiang2023motiongpt} and MotionGPT-2~\cite{wang2024motiongpt2} represent motion through discrete tokens and apply language-model-style sequence learning, while MotionLM~\cite{seff2023motionlm} formulates trajectory forecasting autoregressively.
MotionLLM~\cite{chen2024motionllm} and instruction-tuned motion models~\cite{li2025llamo} further connect motion representations with language, and large-scale efforts such as MoFM~\cite{baharani2025mofm} move toward general-purpose motion foundation models.
A complementary direction learns predictive representations directly in latent space.
I-JEPA~\cite{assran2023ijepa} introduced joint-embedding prediction for images, V-JEPA and V-JEPA~2~\cite{bardes2024vjepa,assran2025vjepa2} extend this principle to temporal visual dynamics, and S-JEPA~\cite{abdelfattah2024sjepa} adapts predictive embedding learning to skeleton sequences.
Related world-model formulations explicitly model future agent dynamics: Driver-WM~\cite{driverwm2026} predicts driver-centered in-cabin dynamics conditioned on traffic information, while risk-aware driver monitoring~\cite{riskaware2026} integrates driver-state world modeling with selective sensing.
More broadly, agent-modeling approaches such as machine theory of mind~\cite{rabinowitz2018tomnet} and mental world modeling~\cite{fei2026mental} emphasize learning representations that predict how agents' internal states and behaviors evolve over time.
These developments create a growing need for long, continuous behavioral sequences that can support token-based, latent-predictive, generative, and multimodal sequence learning.
DriveMotion provides such a testbed in the driver domain, with a common sequence interface that enables comparison across conventional sequence models~\cite{vaswani2017attention,cho2014gru}, generative objectives~\cite{ho2020ddpm,sohn2015cvae}, motion-language models, and predictive world models.

\section{Benchmark protocol details}
\label{sec:app_protocol}

\paragraph{Evaluation subset.} The anchored board (Protocol A) evaluates every model on the identical full set of 16{,}638 anchored windows. For the natural-distribution reference board, the 94{,}642 usable test windows exceed what heavy generative models can be evaluated on at reasonable cost, so it uses a fixed 20{,}000-window subset drawn once with a public seed; best-of-$N$ evaluation uses a smaller fixed subset for the same reason. Both subsets are defined by (sequence id, window start) lists whose hash is checked by the evaluation runner on every invocation, so every number in this paper is computed on byte-identical inputs. The subsets, seeds, and hashes ship with the toolkit.

\paragraph{Evaluation points.} The 23 evaluation points are the nose, eyes, ears, shoulders, and hips plus the 14-point jaw line, the joints that are observable across all camera placements. Errors are computed in the canonical torso frame (origin mid-shoulder, unit median shoulder width) and multiplied by 100 for readability; joints invalid in the ground-truth mask $M$ never contribute.

\paragraph{Head metric masking.} 86.7\% of test-subset windows carry head annotations, but only the BATON DMS-grounded ones (49.8\% of the subset) are used for head-MAE and Motion-F1 in the main table: PnP-derived head pose on web footage has a yaw standard deviation of 0.65\,rad against 0.17\,rad for DMS, and registers ``excursion events'' on 97\% of windows (6.1\% on DMS), a noise floor, not a signal. Vision-label head errors are reported in Table~\ref{tab:head} context for completeness.

\paragraph{Freeze rate.} Freeze rate is the fraction of forecasts whose mean joint net displacement (first to last forecast frame, both ends valid in $M$) is below 0.02 canonical units. The ground-truth frozen fraction under this definition is 0.057; ground-truth median net displacement is 0.078. A per-frame-displacement definition would be confounded by keypoint jitter (ground-truth median 0.028 per frame vs.\ true drift ${\sim}0.003$ per frame) and classify well-smoothed forecasts as frozen.

\paragraph{Point predictions of stochastic models.} The CVAE is scored by its prior-mean decode ($z{=}0$); the DDPM by the mean of 5 samples (a single sample carries irreducible sampling jitter into point metrics); AR-LM and the LLM baseline by one sampled trajectory, with autoregressive decoding over 720 tokens per window dominating their evaluation cost.

\paragraph{CAN event bank.} Events are extracted from synchronized CAN: turns and lane changes from steering angle and net yaw (turn side taken from net yaw when $|{\Delta}\text{yaw}| \geq 5^\circ$, else from mean steering angle), acceleration events from longitudinal speed, hard braking from deceleration thresholds. The regenerated bank contains 76{,}026 events over 1{,}395 routes; turn-left/turn-right counts are balanced (4{,}800 / 4{,}556) and signed magnitudes are consistent with labels by construction.

\paragraph{Anchored window sets.} Protocol A test windows: 6{,}138 pre, 6{,}238 post, 4{,}262 route-matched stable (16{,}638 total), built from test-split routes only after the identity-disjoint split is frozen; windows whose vehicle never exceeds 1\,m/s are excluded from every stratum, stable controls additionally require mean speed ${\geq}2$\,m/s, and per-route stable quotas mirror that route's pre counts. The anchored training-split pool contains 118{,}223 windows (40{,}513 pre / 41{,}498 post / 36{,}212 stable); maneuver-enriched training mixes the pre and post windows (82{,}011) into the natural distribution at the official ${\approx}$48\% rate. Window lists ship as parquet files with the toolkit.

\section{Metric definitions}
\label{sec:app_metrics}

The two primary metrics (MPJPE@4s, Part-State F1@2s) are defined in the main text; this section specifies the full metric suite used in the extended results. The former \emph{Motion-F1} metric is reported here as \textbf{Head-Excursion F1} to make explicit that it is a rare-event head diagnostic (${>}25^\circ$ excursions, 6.1\% base rate on hardware-grounded labels), not a primary behavioral score: regression-trained point forecasters smooth amplitudes below the excursion threshold, so near-zero values diagnose under-motion rather than absence of signal (cf.\ the freeze diagnostics of Sec.~\ref{sec:app_diag}).

On the 23 stable head-and-torso evaluation points, masked by $M$ (units: $10^{-2}$ shoulder widths), we report \textbf{MPJPE@$\tau$} (mean per-joint error up to horizon $\tau \in \{0.5, 1, 2, 4\}$\,s), \textbf{FDE@4s} (error at the final frame), \textbf{head-MAE} (mean absolute head-pose error in degrees), \textbf{Motion-F1} (detection of head excursions ${>}25^\circ$ within the horizon, does the model predict \emph{events} rather than averages?), and \textbf{freeze rate}. The 23-point set contains the joints that are stable across every camera placement and therefore supports fair cross-view comparison; because driver behavior lives disproportionately in the arms, we additionally report \textbf{region MPJPE} over head, torso, and an \emph{action region} (elbows and wrists), so that whole-body claims are checked where they matter behaviorally (Sec.~\ref{sec:app_strata}). Two further metric-design decisions matter on this data. First, \emph{head-derived metrics (head-MAE, Motion-F1) are computed on the DMS-grounded windows only}: vision-derived head pose on web footage is noisy enough that 97\% of web windows would register a ``${>}25^\circ$ excursion'' against a 6.1\% rate on hardware-grounded labels, so scoring against it measures annotation noise, not prediction (vision-label results are reported separately in Sec.~\ref{sec:app_strata}). Second, freeze rate is defined on \emph{net} displacement, mean joint travel from the first to the last forecast frame below $0.02$ canonical units, because per-frame displacement is dominated by keypoint jitter (ground-truth median $0.028$ per frame against a true per-frame drift of ${\sim}0.003$), which would misclassify any well-smoothed forecast as frozen. Under the net definition the ground-truth frozen fraction is $0.057$, the reference against which model collapse should be judged. Stochastic models additionally report \textbf{minADE/minFDE@$N$} and average pairwise diversity (this appendix).

Statistical practice is uniform across both protocols. Overlapping windows from one route are not independent samples, so confidence intervals use a \emph{clustered bootstrap} that resamples routes (Protocol B) or event clusters (Protocol A), never individual windows. Deterministic metrics use a fixed 20{,}000-window test subset drawn once with a public seed and shared verbatim across all models (best-of-$N$: a smaller fixed subset, marked in tables); subset definitions and hashes ship with the toolkit (Sec.~\ref{sec:app_protocol}). Lightweight models report mean$\pm$std over 3 seeds; heavy generative models and Driver-WM$^\dagger$ use one seed, stated explicitly.

\paragraph{Part-level action-state metric.} Coordinate error is the general score, but downstream driver monitoring rarely needs coordinates: it needs to know \emph{whether} the head, torso, or a hand is about to move, and how much. Protocol A therefore also scores each forecast categorically. For each of four parts, head (nose/eyes/ears), torso (shoulders/hips), left and right arm (elbow/wrist), the upcoming motion state within $\tau \in \{1,2\}$\,s is one of \emph{still}, \emph{small}, or \emph{large}; ground-truth states threshold the part's amplitude (masked mean displacement from the last observed frame, maximized over the horizon) at the per-part, per-horizon 50th and 85th percentiles of the training split, frozen and released with the toolkit; a part is excluded from a window when its joint coverage falls below 50\%. Every forecaster is scored by applying the same thresholds to its predicted trajectories at zero additional training; metrics are per-part macro-F1 over the three states and two threshold-free ordinal AUCs (still-vs-moving, large-vs-rest), with a \emph{context carry-over} prior (the part's amplitude over the last $\tau$ seconds of observation, capturing motion persistence) and a linear probe on frozen encoder features as references. This axis captures action anticipation that coordinate error provably misses (the main text (Sec.~5)).

\section{Training details}
\label{sec:app_training}

\paragraph{Baseline families.}
We evaluate every architecture in the context-conditioned configuration and its pose-only ablation, under one training and evaluation harness with matched inputs, parameter budgets, and schedules (Sec.~\ref{sec:app_training}). \textbf{Trivial:} zero-motion (repeat last pose) and visibility-guarded constant velocity. \textbf{Deterministic:} a GRU sequence-to-sequence model~\cite{cho2014gru,martinez2017human}, siMLPe~\cite{guo2023simlpe}, and a Transformer encoder--decoder~\cite{vaswani2017attention} with learned horizon queries, the latter in three sizes (7.6M/44.5M/133M) forming a capability-scaling family. Exterior features enter the Transformer as additional encoder tokens with a learned type embedding, and the recurrent models as a pooled context vector added to the encoder state; missing-scene windows carry the zero-plus-flag convention of the main text (Sec.~4.1). \textbf{Stochastic:} a CVAE~\cite{sohn2015cvae}, a sequence-space diffusion model in the spirit of HumanMAC~\cite{ho2020ddpm,chen2023humanmac}, and a motion-token autoregressive LM in the spirit of MotionGPT/MotionLM~\cite{jiang2023motiongpt,seff2023motionlm} over uniformly quantized core-joint tokens. \textbf{Representation-learning:} a JEPA-style latent predictor~\cite{assran2023ijepa,bardes2024vjepa}, an EMA target encoder provides future latent targets for a horizon-query predictor, with a light decoder head trained on target latents and applied to predicted latents at inference. \textbf{Pretrained LLM:} a frozen Llama-3.2-1B backbone with LoRA adapters~\cite{hu2022lora} and trainable motion-token input/output layers over the same 256-bin quantization as the AR-LM~\cite{jiang2023motiongpt,chen2024motionllm}; the two share everything except the backbone, isolating whether language pretraining transfers to driver motion. A backbone ladder repeats this setup with Llama-3.2-3B (identical adapters, quantization, and schedule; batch size reduced to fit memory), reported as LLM (Llama-3B) in the result tables. \textbf{Driver world model:} no official implementation of Driver-WM~\cite{driverwm2026} is released, so we build a \emph{Driver-WM-style adapted baseline} following its published architecture, dual-stream causal encoding, cross-attention context summaries, a learned vector gate for external-context injection, closed-loop latent rollout, and its skeleton, latent-consistency, and physical-prior losses, with inputs adapted to this benchmark: internal latents come from a learned pose encoder rather than frozen VLM features of cabin video, and the external stream consumes CAN channels rather than exterior video features. The adaptation tests whether a recent driver-centric latent-rollout formulation transfers to pose forecasting; it is not a reproduction of Driver-WM, and we do not compare against its published numbers. All reimplementations are marked ``our impl.''; every learned decoder, including Driver-WM$^\dagger$, uses the same residual convention (predicting displacement from the last observed pose~\cite{martinez2017human}), which we found essential to avoid mean-collapse on this data. Beyond architectures, the benchmark defines two objective-level baselines that target the action-state axis directly: \emph{calibrated decoding} (a per-part monotone amplitude recalibration fitted on anchored training windows) and \emph{reward-time RL fine-tuning} (policy-gradient training of the token model with the part-state metric as reward); both are specified in Sec.~\ref{sec:app_training}.

All models train with AdamW (lr $3{\times}10^{-4}$, weight decay $10^{-4}$), cosine decay with 500-step warmup, gradient clipping at 1.0, bfloat16 autocast, batch size 256 (AR-LM: 32), and 20k steps on the identity-disjoint training split; light models repeat over seeds $\{0,1,2\}$. The shared loss is a masked Huber on future pose plus a velocity-matching term and a 0.2-weighted head MSE; the CVAE adds a KL term annealed to $10^{-2}$ over the first 2{,}000 steps; the DDPM trains $\epsilon$-prediction over 100 diffusion steps on anchor-relative futures; the AR-LM trains next-token cross-entropy over 256-bin uniformly quantized core-joint tokens; Driver-WM$^\dagger$ optimizes its published composite loss (skeleton, latent velocity consistency, bone length, smoothness) with the latent targets produced by causally encoding the ground-truth future with the same pre-encoder. Parameter counts are 1.8M (GRU), 1.8M (siMLPe), 7.6M (Transformer), 2.3M (CVAE), 12.8M (DDPM), 14.4M (AR-LM), 5.0M (Driver-WM$^\dagger$). All decoders predict displacement from the last observed pose~\cite{martinez2017human}; ablating this residual anchor caused every architecture to collapse toward a mean pose (e.g., Driver-WM$^\dagger$ validation error degrades from 0.213 to 0.666 canonical units).

\paragraph{RL fine-tuning of the token model.} The reward-time baseline fine-tunes the trained AR-LM with a group-relative policy gradient (GRPO-style): for each maneuver-anchored training window, $K{=}4$ futures are sampled, each is scored by its part-state agreement with ground truth (the Protocol A action-state metric), and the advantage of each sample against the group mean weights the exact token log-probabilities of that sample; a cross-entropy anchor to the data (weight 0.5) prevents policy drift. Training uses batch 8, lr $10^{-5}$, on pre/post training windows only, with a 1{,}500-step budget; the run was stopped after roughly 620 steps once the reward showed no trend (Sec.~\ref{sec:app_diag}). This optimizes the benchmark's state metric directly; coordinate metrics are monitored, not optimized.

\section{Natural-distribution reference board}
\label{sec:app_natboard}

\begin{table*}[t]
\centering
\caption{Natural-distribution reference board (fixed 20k-window subset); ground-truth freeze rate 0.057.}
\label{tab:main}
\small
\setlength{\tabcolsep}{4.5pt}
\adjustbox{max width=\textwidth}{%
\rowcolors{3}{white}{dmblue3}
\begin{tabular}{@{\extracolsep{\fill}}lcccccccc}
\toprule
 & \multicolumn{4}{c}{MPJPE $\downarrow$} & FDE@4s $\downarrow$ & head-MAE@4s $\downarrow$ & Head-Exc.\ F1 $\uparrow$ & Freeze \\
Model & @0.5s & @1s & @2s & @4s & & & & rate \\
\midrule
Zero-motion & 5.5 & 6.9 & 8.5 & 10.2 & 12.7 & 4.8 & 0.00 & 1.00 \\
Constant-velocity & 12.6 & 21.4 & 38.5 & 72.2 & 137.7 & 4.8 & 0.00 & 0.01 \\
\midrule
GRU seq2seq (our impl.) & 5.9{\scriptsize$\pm$0.0} & 6.9{\scriptsize$\pm$0.0} & 8.1{\scriptsize$\pm$0.0} & 9.5{\scriptsize$\pm$0.0} & 11.5{\scriptsize$\pm$0.1} & 4.2{\scriptsize$\pm$0.0} & 0.07{\scriptsize$\pm$0.01} & 0.22{\scriptsize$\pm$0.01} \\
siMLPe~\cite{guo2023simlpe} (our impl.) & 5.8{\scriptsize$\pm$0.0} & 6.9{\scriptsize$\pm$0.0} & 8.1{\scriptsize$\pm$0.0} & 9.5{\scriptsize$\pm$0.0} & 11.5{\scriptsize$\pm$0.0} & 4.4{\scriptsize$\pm$0.0} & 0.00{\scriptsize$\pm$0.00} & 0.33{\scriptsize$\pm$0.01} \\
Transformer ED (our impl.) & 5.3{\scriptsize$\pm$0.0} & 6.4{\scriptsize$\pm$0.0} & 7.7{\scriptsize$\pm$0.0} & 9.0{\scriptsize$\pm$0.0} & 10.9{\scriptsize$\pm$0.0} & 4.2{\scriptsize$\pm$0.0} & 0.09{\scriptsize$\pm$0.02} & 0.51{\scriptsize$\pm$0.01} \\
\midrule
CVAE (our impl.) & 5.6{\scriptsize$\pm$0.0} & 6.6{\scriptsize$\pm$0.0} & 7.9{\scriptsize$\pm$0.0} & 9.4{\scriptsize$\pm$0.0} & 11.4{\scriptsize$\pm$0.0} & 4.2{\scriptsize$\pm$0.0} & 0.05{\scriptsize$\pm$0.00} & 0.34{\scriptsize$\pm$0.01} \\
DDPM (our impl.) & 7.2 & 8.4 & 9.7 & 11.3 & 13.7 & 6.4 & 0.06 & 0.00 \\
Motion-token AR-LM (our impl.)\textsuperscript{s} & 7.9 & 9.2 & 10.8 & 12.7 & 15.3 & 5.9 & 0.00 & 0.62 \\
\midrule
JEPA latent predictor~\cite{assran2023ijepa} (our impl.) & 5.8 & 7.1 & 8.6 & 10.3 & 12.7 & 4.7 & 0.01 & 0.39 \\
Driver-WM$^\dagger$~\cite{driverwm2026} (adapted, our impl.) & 5.6 & 7.0 & 8.6 & 10.4 & 13.0 & 4.8 & 0.00 & 0.83 \\
\midrule
LLM (Llama-3.2-1B + LoRA)~\cite{chen2024motionllm} (our impl.)\textsuperscript{s} & 8.0 & 9.3 & 11.0 & 13.0 & 15.9 & 5.9 & 0.00 & 0.55 \\
\bottomrule
\end{tabular}}
\end{table*}

Table~\ref{tab:main} shows what happens when driver motion is evaluated the way generic human-motion benchmarks are: on uniformly sampled windows. Repeating the last pose achieves 10.2 MPJPE@4s; the best learned model reaches 9.0, a real but modest 12\% margin, while extrapolating instantaneous velocity amplifies keypoint jitter into a $7\times$ larger error (Sec.~\ref{sec:app_diag}). Worse, low average error does not mean the motion that matters is forecast: against the ground-truth references (frozen fraction 0.057, head-event rate 6.1\%), every learned point forecaster under-predicts motion amplitude, freeze rates span 0.22--0.51, and even the best model detects only a small fraction of upcoming large head excursions (Motion-F1 $\leq$ 0.09). A model can score well on aggregate MPJPE while failing to anticipate precisely the sparse, behaviorally meaningful movements a monitoring system needs.

Stratifying the same test set by ground-truth motion makes the structure explicit (the motion-active table in the main text): on the quiet half every model ties with persistence, as it should, while on the motion-active quartile the Transformer's absolute margin over zero-motion doubles (19.91${\to}$17.45, $-2.46$ vs.\ $-1.20$ overall). The learned margin is therefore concentrated in the moments when the driver acts, which motivates an evaluation that resolves this structure instead of averaging over it. Protocol A anchors evaluation on the vehicle-dynamics transitions around which motion-active moments cluster.

Motion-active stratification (quiet half vs.\ active quartile) is reported in the main text (the motion-active table in the main text).

\section{Full anchored board and ablations}
\label{sec:app_fullboard}

\paragraph{The premise holds.} Before any model is scored, the anchoring premise itself is measured on ground truth: over the anchored test windows (route-matched, parked-excluded), arm amplitude in pre-maneuver windows is $3.4\times$ that of stable-driving controls, and mean ground-truth motion in the pre and post strata exceeds the stable stratum by $1.7\times$ in MPJPE terms (zero-motion: 8.85/9.07 vs.\ 5.33). The intervals Protocol A selects are, verifiably, where driver behavior lives, and because they are selected by \emph{vehicle} state, the selection never inspects the driver-motion target itself.

\begin{table*}[t]
\centering
\caption{Full anchored board. All models except the RL fine-tuned variant are evaluated on the identical 16{,}638 anchored windows; the RL row uses a reduced subset. \textsuperscript{p} = pose-only.}
\label{tab:dynfull}
\small
\setlength{\tabcolsep}{4.5pt}
\adjustbox{max width=\textwidth}{%
\rowcolors{3}{white}{dmblue3}
\begin{tabular}{@{\extracolsep{\fill}}lcccccccc}
\toprule
 & \multicolumn{4}{c}{MPJPE@4s $\downarrow$} & \multicolumn{4}{c}{State-F1@2s $\uparrow$} \\
Model & Pre & Post & Stable & Macro & Pre & Post & Stable & Macro \\
\midrule
Zero-motion & 8.85 & 9.07 & 5.33 & 7.75 & 0.196 & 0.180 & 0.269 & 0.215 \\
\midrule
GRU seq2seq & 7.84{\scriptsize$\pm$0.01} & 8.00{\scriptsize$\pm$0.01} & 4.67{\scriptsize$\pm$0.01} & 6.83{\scriptsize$\pm$0.01} & 0.259{\scriptsize$\pm$0.003} & 0.245{\scriptsize$\pm$0.002} & 0.322{\scriptsize$\pm$0.001} & 0.275{\scriptsize$\pm$0.001} \\
siMLPe\textsuperscript{p}~\cite{guo2023simlpe} & 7.82{\scriptsize$\pm$0.01} & 8.07{\scriptsize$\pm$0.00} & 4.66{\scriptsize$\pm$0.00} & 6.85{\scriptsize$\pm$0.00} & 0.209{\scriptsize$\pm$0.001} & 0.194{\scriptsize$\pm$0.001} & 0.278{\scriptsize$\pm$0.001} & 0.227{\scriptsize$\pm$0.001} \\
Transformer ED & 7.76{\scriptsize$\pm$0.01} & 7.94{\scriptsize$\pm$0.02} & 4.56{\scriptsize$\pm$0.01} & 6.75{\scriptsize$\pm$0.01} & 0.265{\scriptsize$\pm$0.001} & 0.252{\scriptsize$\pm$0.003} & 0.328{\scriptsize$\pm$0.002} & 0.282{\scriptsize$\pm$0.002} \\
Transformer ED (enriched) & 7.99{\scriptsize$\pm$0.05} & 8.18{\scriptsize$\pm$0.04} & 4.68{\scriptsize$\pm$0.01} & 6.95{\scriptsize$\pm$0.03} & 0.299{\scriptsize$\pm$0.002} & 0.289{\scriptsize$\pm$0.001} & 0.338{\scriptsize$\pm$0.003} & 0.309{\scriptsize$\pm$0.001} \\
Transformer-L\textsuperscript{p} & 7.63{\scriptsize$\pm$0.00} & 7.82{\scriptsize$\pm$0.00} & 4.44{\scriptsize$\pm$0.01} & 6.63{\scriptsize$\pm$0.00} & 0.269{\scriptsize$\pm$0.000} & 0.256{\scriptsize$\pm$0.001} & 0.335{\scriptsize$\pm$0.001} & 0.287{\scriptsize$\pm$0.001} \\
Transformer-XL\textsuperscript{p} & 7.64 & 7.81 & 4.42 & 6.62 & 0.266 & 0.258 & 0.329 & 0.284 \\
\midrule
CVAE & 7.85{\scriptsize$\pm$0.01} & 8.01{\scriptsize$\pm$0.01} & 4.65{\scriptsize$\pm$0.01} & 6.84{\scriptsize$\pm$0.01} & 0.240{\scriptsize$\pm$0.002} & 0.227{\scriptsize$\pm$0.002} & 0.303{\scriptsize$\pm$0.001} & 0.257{\scriptsize$\pm$0.002} \\
DDPM\textsuperscript{p} & 9.88 & 10.07 & 6.78 & 8.91 & 0.287 & 0.280 & 0.326 & 0.298 \\
DDPM\textsuperscript{p} (enriched) & 9.88 & 10.06 & 6.79 & 8.91 & 0.292 & 0.277 & 0.324 & 0.298 \\
DDPM & 9.89 & 10.04 & 6.80 & 8.91 & 0.290 & 0.278 & 0.324 & 0.297 \\
DDPM (enriched) & 9.83{\scriptsize$\pm$0.00} & 10.00{\scriptsize$\pm$0.01} & 6.75{\scriptsize$\pm$0.01} & 8.86{\scriptsize$\pm$0.00} & 0.292{\scriptsize$\pm$0.002} & 0.283{\scriptsize$\pm$0.003} & 0.323{\scriptsize$\pm$0.001} & 0.299{\scriptsize$\pm$0.001} \\
AR-LM\textsuperscript{p}~\cite{jiang2023motiongpt} & 9.25 & 9.49 & 5.67 & 8.14 & 0.452 & 0.442 & 0.462 & 0.452 \\
AR-LM\textsuperscript{p} (enriched) & 9.28 & 9.52 & 5.68 & 8.16 & 0.458 & 0.454 & 0.461 & 0.458 \\
AR-LM (enriched) & 9.23 & 9.48 & 5.66 & 8.12 & 0.452 & 0.447 & 0.456 & 0.452 \\
AR-LM\textsuperscript{p} (RL fine-tuned) & 9.29 & 9.71 & 6.18 & 8.39 & 0.465 & 0.435 & 0.451 & 0.450 \\
LLM (Llama-1B)\textsuperscript{p} & 9.47 & 9.71 & 5.79 & 8.33 & 0.451 & 0.450 & 0.456 & 0.452 \\
LLM (Llama-3B)\textsuperscript{p} & 9.34 & 9.63 & 5.74 & 8.24 & 0.456 & 0.451 & 0.464 & 0.457 \\
LLM (Qwen-VL-3B)\textsuperscript{p} & 9.36 & 9.61 & 5.75 & 8.24 & 0.459 & 0.455 & 0.458 & 0.457 \\
\bottomrule
\end{tabular}}
\end{table*}

\paragraph{Reading the board.} On the anchored windows of Table~\ref{tab:dynfull}, learned forecasting improves over persistence by 13--15\% (macro MPJPE@4s 7.75${\to}$6.75 for the base Transformer, 6.63/6.62 for L/XL), a larger margin than the 12\% observed under uniform sampling. Several regularities hold across architectures. The strata order consistently for every model: pre and post are harder than stable in absolute error. Coordinate error and Part-State F1 give different rankings. The motion-token models have substantially higher coordinate error than the regression models (8.12--8.33 macro) and the highest raw Part-State F1 (0.452 vs.\ 0.309 for the best regression model). Three token variants (task-trained, context-plus-enrichment, and the Llama-1B backbone) reach an identical 0.452 macro state-F1, and enriched pose-only training reaches 0.458, so the raw state advantage is a property of discrete decoding itself, with backbone, conditioning, and enrichment moving it by at most $0.006$; on the coordinate axis, context and enrichment improve token models from 8.16 to 8.12 macro. Scaling the Transformer from 7.6M to 44.5M parameters improves every stratum at negligible seed variance (pose-only: 7.73${\to}$7.63 pre, 4.50${\to}$4.44 stable), and a further step to 133M holds the coordinate values without extending them (7.64 pre, 4.42 stable; 1 seed).

The context-vs-pose-only pairing is tabulated in Table~\ref{tab:sceneabl}.

\begin{table*}[htbp]
\centering
\caption{Exterior context vs.\ pose-only; $\Delta$ = macro MPJPE@4s change.}
\label{tab:sceneabl}
\small
\setlength{\tabcolsep}{8pt}
\adjustbox{max width=\textwidth}{%
\rowcolors{3}{white}{dmblue3}
\begin{tabular}{@{\extracolsep{\fill}}lccccccc}
\toprule
 & \multicolumn{3}{c}{Macro@4s} & \multicolumn{2}{c}{Pre@4s} & \multicolumn{2}{c}{State-F1} \\
Model & pose & +ctx & $\Delta$ & pose & +ctx & pose & +ctx \\
\midrule
Transformer ED & 6.71{\scriptsize$\pm$0.01} & 6.75{\scriptsize$\pm$0.01} & +0.7\% & 7.73{\scriptsize$\pm$0.01} & 7.76{\scriptsize$\pm$0.01} & 0.276{\scriptsize$\pm$0.001} & 0.282{\scriptsize$\pm$0.002} \\
Transformer ED (enriched) & 6.74{\scriptsize$\pm$0.01} & 6.95{\scriptsize$\pm$0.03} & +3.1\% & 7.75{\scriptsize$\pm$0.01} & 7.99{\scriptsize$\pm$0.05} & 0.292{\scriptsize$\pm$0.001} & 0.309{\scriptsize$\pm$0.001} \\
GRU seq2seq & 6.81{\scriptsize$\pm$0.01} & 6.83{\scriptsize$\pm$0.01} & +0.3\% & 7.83{\scriptsize$\pm$0.01} & 7.84{\scriptsize$\pm$0.01} & 0.277{\scriptsize$\pm$0.004} & 0.275{\scriptsize$\pm$0.001} \\
CVAE & 6.82{\scriptsize$\pm$0.00} & 6.84{\scriptsize$\pm$0.01} & +0.3\% & 7.84{\scriptsize$\pm$0.01} & 7.85{\scriptsize$\pm$0.01} & 0.257{\scriptsize$\pm$0.002} & 0.257{\scriptsize$\pm$0.002} \\
\bottomrule
\end{tabular}}
\end{table*}

Pairing each architecture with its pose-only ablation under identical training (Table~\ref{tab:sceneabl}) isolates the effect of the exterior stream. With the frozen, mean-pooled features of the reference implementation, macro MPJPE@4s differs by less than 1\% in either direction across architectures. Part-State F1 improves in the pre and post strata on the Transformer (pre 0.259${\to}$0.265, post 0.247${\to}$0.252, both $\pm 0.001$--$0.003$ over 3 seeds), and on the natural distribution the freeze rate drops from 0.51 to 0.40. Relative deltas are similar on windows with and without exterior coverage (has\_scene stratification), so the gain is not yet driven by fine-grained scene content.

\paragraph{Context and enrichment combine.} Under maneuver-enriched training, exterior context lifts pre-stratum Part-State F1 from 0.275 to 0.299 and macro F1 from 0.291 to 0.309 ($\pm 0.001$--$0.002$ over 3 seeds), the highest regression-model value on the board and above the diffusion model (0.298); the same context added to natural-distribution training is worth only $+0.006$. The combined gain over the plain pose-only model ($+0.040$ pre-stratum F1) is roughly double the sum of the two individual gains, consistent with enrichment exposing the model to the maneuver-coupled intervals in which the exterior stream is informative. Scene-plus-enrichment at the official 48\% dose also matches the state anticipation of the 90\%-dose pose-only model at slightly lower coordinate cost. These numbers characterize the reference features, a frozen ImageNet encoder pooled to one vector per half-second; the full synchronized streams are released so that stronger scene encoders can be evaluated against these bounds.
Token models behave differently: adding context and enrichment to the AR-LM leaves Part-State F1 unchanged (0.452 in both configurations) but improves coordinates (8.14${\to}$8.12 macro), the opposite pattern to the regression families, whose training-time levers change states and leave geometry unchanged.

Protocol A prescribes maneuver-enriched training; the prescription is validated by dose. Holding the pose-only Transformer fixed and raising the fraction of anchored windows in the training mix from 0\% to 48\% to 90\% moves Part-State F1 monotonically, pre-stratum 0.259${\to}$0.275${\to}$0.298, macro 0.275${\to}$0.291${\to}$0.309, while coordinate error is flat at the official dose (macro MPJPE@4s 6.70${\to}$6.74) and mildly degraded only at 90\% (6.99, $+4\%$): at the official dose, enrichment changes which motion is produced without a coordinate cost. The gains not only persist but grow in the context-conditioned configuration (see the context results above). The official mix (${\approx}$48\%) is chosen on this curve: it captures most of the state-anticipation gain at zero coordinate cost, and retains exposure to the natural distribution; Protocol B indicates that this exposure supports cross-domain transfer.

\section{Calibrated decoding}
\label{sec:app_calib}

\begin{table}[t]
\centering
\caption{Part-State F1 before/after amplitude recalibration.}
\label{tab:calib}
\small
\setlength{\tabcolsep}{3.5pt}
\adjustbox{max width=\linewidth}{%
\rowcolors{2}{white}{dmblue3}
\begin{tabular}{@{\extracolsep{\fill}}lccccc}
\toprule
 & \multicolumn{2}{c}{raw} & \multicolumn{2}{c}{calibrated} & \\
Model & Pre & Macro & Pre & Macro & $\Delta$ \\
\midrule
Transformer ED\textsuperscript{p} & 0.259{\scriptsize$\pm$0.001} & 0.276{\scriptsize$\pm$0.001} & 0.522{\scriptsize$\pm$0.002} & 0.514{\scriptsize$\pm$0.002} & +0.237 \\
Transformer-L\textsuperscript{p} & 0.269{\scriptsize$\pm$0.000} & 0.287{\scriptsize$\pm$0.001} & 0.529 & 0.517 & +0.230 \\
Transformer ED (+ctx) & 0.265{\scriptsize$\pm$0.001} & 0.282{\scriptsize$\pm$0.002} & 0.518{\scriptsize$\pm$0.004} & 0.507{\scriptsize$\pm$0.003} & +0.225 \\
Transformer ED (+ctx, enriched) & 0.299{\scriptsize$\pm$0.002} & 0.309{\scriptsize$\pm$0.001} & 0.492{\scriptsize$\pm$0.004} & 0.486{\scriptsize$\pm$0.002} & +0.177 \\
DDPM (+ctx, enriched) & 0.292{\scriptsize$\pm$0.002} & 0.299{\scriptsize$\pm$0.001} & 0.383{\scriptsize$\pm$0.004} & 0.371{\scriptsize$\pm$0.002} & +0.072 \\
\bottomrule
\end{tabular}}
\end{table}

\paragraph{Effect of calibration.} A per-part monotone amplitude recalibration, fitted only on anchored training windows, raises macro Part-State F1 from 0.28--0.31 to 0.49--0.52 across every Transformer configuration ($+0.18$ to $+0.24$ absolute), with coordinate metrics unchanged by construction: the trajectory coordinates are not modified, and the calibration applies only to the per-part amplitude used for state decoding (Table~\ref{tab:calib}). The best calibrated configuration, Transformer-L, reaches 0.517 macro F1, compared with 0.59--0.65 per part for the carry-over prior computed from observed ground truth. Calibration also changes the ordering: after post-hoc amplitude correction, performance is determined by ranking quality, where the natural-distribution pose-only models lead (0.513--0.517) and the enriched-scene model retains less headroom (0.486), since its training-time gains already came from producing better amplitudes. Raw forecast-derived states and calibrated states therefore measure different quantities, and the benchmark reports both.
A reward-time route, GRPO-style RL fine-tuning of the token model with the Part-State metric as reward (Sec.~\ref{sec:app_training}), was stopped after roughly 620 of a budgeted 1{,}500 steps without measurable gain (reward flat at ${\sim}$0.45 throughout; the evaluated mid-training checkpoint, step ${\sim}$350, scores 0.450 vs.\ the 0.452 baseline); the group-relative reward signal is dominated by window-difficulty variance at this scale, and we report it as an open direction rather than a working lever.

\section{Part-state anticipation details}
\label{sec:app_state}

The question a monitoring system actually asks is rarely \emph{where} the wrist will be at 4\,s, it is whether a hand is about to move, whether the head will turn, whether motion is about to grow. The action-state metric evaluates exactly this question, and Table~\ref{tab:state} reports it on the natural distribution, where three structural results emerge. First, \emph{motion persistence carries most of the predictable signal}: the context carry-over prior (F1 0.59--0.65) and the linear probe on frozen encoder features (0.52--0.58) both far exceed every forecast-derived predictor, trajectory forecasters know more about upcoming motion states than their own decoded trajectories express. Second, the forecast-derived ordering is not the MPJPE ordering: token-based models (AR-LM, LLM) are weak coordinate regressors but the strongest forecast-derived state predictors on torso and arms (F1 up to 0.57), because discrete decoding preserves motion magnitude that regression smooths away. Third, ordinal AUCs (0.7--0.9) show all learned forecasters rank upcoming motion far better than their thresholded states suggest, the gap between AUC and F1 is decoder miscalibration, and it is directly fixable.

\begin{table*}[t]
\centering
\caption{Per-part state anticipation: macro-F1 and still-vs-moving AUC.}
\label{tab:state}
\small
\setlength{\tabcolsep}{4pt}
\adjustbox{max width=\textwidth}{%
\rowcolors{3}{white}{dmblue3}
\begin{tabular}{@{\extracolsep{\fill}}lcccccccc}
\toprule
 & \multicolumn{2}{c}{Head} & \multicolumn{2}{c}{Torso} & \multicolumn{2}{c}{L-arm} & \multicolumn{2}{c}{R-arm} \\
Method & F1 & AUC & F1 & AUC & F1 & AUC & F1 & AUC \\
\midrule
Context carry-over & 0.61 & 0.81 & 0.65 & 0.86 & 0.59 & 0.82 & 0.64 & 0.88 \\
\midrule
Zero-motion (derived) & 0.20{\scriptsize$\pm$0.00} & 0.32{\scriptsize$\pm$0.01} & 0.18{\scriptsize$\pm$0.01} & 0.27{\scriptsize$\pm$0.00} & 0.19{\scriptsize$\pm$0.00} & 0.39{\scriptsize$\pm$0.00} & 0.17{\scriptsize$\pm$0.00} & 0.24{\scriptsize$\pm$0.01} \\
\midrule
GRU seq2seq & 0.28{\scriptsize$\pm$0.00} & 0.78{\scriptsize$\pm$0.01} & 0.18{\scriptsize$\pm$0.00} & 0.76{\scriptsize$\pm$0.00} & 0.24{\scriptsize$\pm$0.00} & 0.73{\scriptsize$\pm$0.00} & 0.27{\scriptsize$\pm$0.01} & 0.81{\scriptsize$\pm$0.01} \\
siMLPe & 0.23{\scriptsize$\pm$0.00} & 0.75{\scriptsize$\pm$0.00} & 0.23{\scriptsize$\pm$0.01} & 0.76{\scriptsize$\pm$0.01} & 0.22{\scriptsize$\pm$0.00} & 0.73{\scriptsize$\pm$0.00} & 0.21{\scriptsize$\pm$0.00} & 0.80{\scriptsize$\pm$0.00} \\
Transformer ED & 0.26{\scriptsize$\pm$0.00} & 0.75{\scriptsize$\pm$0.01} & 0.17{\scriptsize$\pm$0.00} & 0.71{\scriptsize$\pm$0.02} & 0.22{\scriptsize$\pm$0.00} & 0.75{\scriptsize$\pm$0.00} & 0.23{\scriptsize$\pm$0.00} & 0.82{\scriptsize$\pm$0.01} \\
\midrule
CVAE & 0.26{\scriptsize$\pm$0.01} & 0.77{\scriptsize$\pm$0.00} & 0.17{\scriptsize$\pm$0.00} & 0.74{\scriptsize$\pm$0.01} & 0.22{\scriptsize$\pm$0.00} & 0.72{\scriptsize$\pm$0.01} & 0.22{\scriptsize$\pm$0.02} & 0.79{\scriptsize$\pm$0.01} \\
DDPM & 0.24 & 0.58 & 0.25 & 0.56 & 0.28 & 0.60 & 0.34 & 0.62 \\
AR-LM & 0.26 & 0.67 & 0.55 & 0.79 & 0.48 & 0.73 & 0.57 & 0.84 \\
LLM (Llama-1B) & 0.27 & 0.65 & 0.53 & 0.77 & 0.47 & 0.71 & 0.57 & 0.82 \\
\midrule
JEPA & 0.23 & 0.74 & 0.17 & 0.75 & 0.31 & 0.68 & 0.32 & 0.82 \\
Driver-WM$^\dagger$ & 0.20 & 0.69 & 0.18 & 0.72 & 0.19 & 0.59 & 0.21 & 0.76 \\
\midrule
Linear probe (frozen enc.) & 0.58 & 0.78 & 0.58 & 0.81 & 0.52 & 0.75 & 0.55 & 0.85 \\
\bottomrule
\end{tabular}}
\end{table*}

Per-part state thresholds (canonical units, train-split percentiles) ship in \texttt{state\_thresholds.json}; the label builder, coverage rule (part joint coverage $\geq 50\%$ over the horizon), and the exact part-to-joint assignment are part of the toolkit. The $\tau{=}1$\,s table, per-part Brier scores of the sample-derived probabilities, and the ground-truth oracle check (applying thresholds to ground-truth amplitudes reproduces the labels near-perfectly) are provided with the released evaluation logs.

\section{Stratified results}
\label{sec:app_strata}

\begin{table}[t]
\centering
\caption{Cross-view breakdown (B-view). MPJPE@4s on web test windows by annotated camera view (all-source training). Multi-seed models: mean{\scriptsize$\pm$}std over 3 seeds.}
\label{tab:view}
\small
\setlength{\tabcolsep}{5pt}
\adjustbox{max width=\linewidth}{%
\rowcolors{2}{dmblue3}{white}
\begin{tabular}{@{\extracolsep{\fill}}lccc}
\toprule
Model & Back & Front & Side \\
\midrule
Zero-motion & 11.0 & 19.0 & 13.4 \\
Constant-velocity & 72.9 & 72.4 & 98.3 \\
\midrule
GRU seq2seq (our impl.) & 10.5{\scriptsize$\pm$0.1} & 16.3{\scriptsize$\pm$0.2} & 12.9{\scriptsize$\pm$0.1} \\
siMLPe~\cite{guo2023simlpe} (our impl.) & 10.4{\scriptsize$\pm$0.0} & 18.3{\scriptsize$\pm$0.1} & 12.7{\scriptsize$\pm$0.1} \\
Transformer ED (our impl.) & 9.7{\scriptsize$\pm$0.0} & 15.4{\scriptsize$\pm$0.1} & 12.1{\scriptsize$\pm$0.0} \\
\midrule
CVAE (our impl.) & 10.3{\scriptsize$\pm$0.1} & 16.2{\scriptsize$\pm$0.0} & 12.6{\scriptsize$\pm$0.0} \\
DDPM (our impl.) & 11.9 & 19.9 & 14.2 \\
Motion-token AR-LM (our impl.) & 16.4 & 21.6 & 18.8 \\
\midrule
JEPA latent predictor~\cite{assran2023ijepa} (our impl.) & 11.3 & 19.5 & 13.7 \\
Driver-WM$^\dagger$~\cite{driverwm2026} (adapted, our impl.) & 11.4 & 19.3 & 13.7 \\
\midrule
LLM (Llama-3.2-1B + LoRA)~\cite{chen2024motionllm} (our impl.) & 16.5 & 22.9 & 19.3 \\
\bottomrule
\end{tabular}}
\end{table}

\begin{table}[t]
\centering
\caption{Visibility strata (B-vis). MPJPE@4s on web test windows by annotated driver-visibility level (\emph{arms-only} appears as \texttt{no\_body} in the release keys). Strata also differ in viewpoint and content; numbers are descriptive.}
\label{tab:vis}
\small
\setlength{\tabcolsep}{4pt}
\adjustbox{max width=\linewidth}{%
\rowcolors{2}{dmblue3}{white}
\begin{tabular}{@{\extracolsep{\fill}}lccccc}
\toprule
Model & Body & Eye & Face & Head & Arms-only \\
\midrule
Zero-motion & 30.6 & 13.1 & 12.5 & 36.4 & 36.4 \\
Constant-velocity & 149.4 & 84.6 & 93.8 & 268.4 & 239.6 \\
\midrule
GRU seq2seq (our impl.) & 30.1{\scriptsize$\pm$0.4} & 12.3{\scriptsize$\pm$0.1} & 12.1{\scriptsize$\pm$0.1} & 36.3{\scriptsize$\pm$0.4} & 37.4{\scriptsize$\pm$0.6} \\
siMLPe~\cite{guo2023simlpe} (our impl.) & 28.9{\scriptsize$\pm$0.2} & 12.6{\scriptsize$\pm$0.1} & 11.7{\scriptsize$\pm$0.1} & 34.7{\scriptsize$\pm$0.2} & 35.9{\scriptsize$\pm$0.1} \\
Transformer ED (our impl.) & 28.2{\scriptsize$\pm$0.2} & 11.5{\scriptsize$\pm$0.0} & 11.2{\scriptsize$\pm$0.0} & 35.5{\scriptsize$\pm$0.0} & 36.5{\scriptsize$\pm$0.1} \\
\midrule
CVAE (our impl.) & 30.2{\scriptsize$\pm$0.0} & 12.0{\scriptsize$\pm$0.0} & 11.8{\scriptsize$\pm$0.0} & 36.3{\scriptsize$\pm$0.0} & 37.0{\scriptsize$\pm$0.0} \\
DDPM (our impl.) & 32.0 & 14.0 & 13.3 & 36.8 & 36.5 \\
Motion-token AR-LM (our impl.) & 41.0 & 16.2 & 16.1 & 89.5 & 25.9 \\
\midrule
JEPA latent predictor~\cite{assran2023ijepa} (our impl.) & 30.7 & 13.5 & 12.8 & 37.0 & 36.6 \\
Driver-WM$^\dagger$~\cite{driverwm2026} (adapted, our impl.) & 32.7 & 13.5 & 12.8 & 36.9 & 38.1 \\
\midrule
LLM (Llama-3.2-1B + LoRA)~\cite{chen2024motionllm} (our impl.) & 46.8 & 16.8 & 16.5 & 89.3 & 35.3 \\
\bottomrule
\end{tabular}}
\end{table}

\begin{table}[t]
\centering
\caption{Head-pose fidelity on DMS-grounded labels (B-head). Head-MAE (degrees) on the BATON test windows whose head pose comes from the production DMS stack (9{,}963 windows).}
\label{tab:head}
\small
\setlength{\tabcolsep}{6pt}
\adjustbox{max width=\linewidth}{%
\rowcolors{2}{dmblue3}{white}
\begin{tabular}{@{\extracolsep{\fill}}lcccc}
\toprule
Model & @0.5s & @1s & @2s & @4s \\
\midrule
Zero-motion & 2.21 & 2.99 & 3.88 & 4.80 \\
Constant-velocity & 2.21 & 2.99 & 3.88 & 4.80 \\
GRU seq2seq & 2.24{\scriptsize$\pm$0.01} & 2.89{\scriptsize$\pm$0.01} & 3.58{\scriptsize$\pm$0.01} & 4.21{\scriptsize$\pm$0.01} \\
siMLPe & 2.43{\scriptsize$\pm$0.01} & 3.08{\scriptsize$\pm$0.01} & 3.75{\scriptsize$\pm$0.01} & 4.38{\scriptsize$\pm$0.00} \\
Transformer ED & 2.15{\scriptsize$\pm$0.00} & 2.83{\scriptsize$\pm$0.00} & 3.54{\scriptsize$\pm$0.00} & 4.19{\scriptsize$\pm$0.00} \\
CVAE & 2.20{\scriptsize$\pm$0.00} & 2.87{\scriptsize$\pm$0.00} & 3.57{\scriptsize$\pm$0.00} & 4.21{\scriptsize$\pm$0.00} \\
DDPM & 8.45 & 8.79 & 9.20 & 9.70 \\
Driver-WM$^\dagger$ & 2.27 & 3.04 & 3.93 & 4.83 \\
\bottomrule
\end{tabular}}
\end{table}

\begin{table}[t]
\centering
\caption{Region MPJPE@4s on the main test subset: head (nose/eyes/ears + jaw), torso (shoulders/hips), and the action region (elbows + wrists), alongside the 23-point stable set. The action region carries the largest motion and the largest errors, quantifying the headroom the stable-set metric alone would hide.}
\label{tab:region}
\small
\setlength{\tabcolsep}{5pt}
\adjustbox{max width=\linewidth}{%
\rowcolors{2}{dmblue3}{white}
\begin{tabular}{@{\extracolsep{\fill}}lcccc}
\toprule
Model & Head & Torso & Arms (action) & Stable-23 \\
\midrule
Zero-motion & 10.1 & 10.7 & 28.1 & 10.2 \\
Constant-velocity & 67.5 & 94.3 & 145.5 & 72.2 \\
\midrule
GRU seq2seq (our impl.) & 9.3{\scriptsize$\pm$0.0} & 10.6{\scriptsize$\pm$0.0} & 27.4{\scriptsize$\pm$0.0} & 9.5{\scriptsize$\pm$0.0} \\
siMLPe~\cite{guo2023simlpe} (our impl.) & 9.4{\scriptsize$\pm$0.0} & 10.2{\scriptsize$\pm$0.1} & 27.1{\scriptsize$\pm$0.0} & 9.5{\scriptsize$\pm$0.0} \\
Transformer ED (our impl.) & 8.7{\scriptsize$\pm$0.0} & 10.4{\scriptsize$\pm$0.0} & 26.8{\scriptsize$\pm$0.0} & 9.0{\scriptsize$\pm$0.0} \\
\midrule
CVAE (our impl.) & 9.1{\scriptsize$\pm$0.0} & 10.6{\scriptsize$\pm$0.0} & 27.4{\scriptsize$\pm$0.1} & 9.4{\scriptsize$\pm$0.0} \\
DDPM (our impl.) & 11.1 & 12.4 & 31.1 & 11.3 \\
Motion-token AR-LM (our impl.) & 10.9 & 21.1 & 44.4 & 12.7 \\
\midrule
JEPA latent predictor~\cite{assran2023ijepa} (our impl.) & 10.2 & 10.8 & 27.9 & 10.3 \\
Driver-WM$^\dagger$~\cite{driverwm2026} (adapted, our impl.) & 10.1 & 11.6 & 28.4 & 10.4 \\
\midrule
LLM (Llama-3.2-1B + LoRA)~\cite{chen2024motionllm} (our impl.) & 11.0 & 22.4 & 47.9 & 13.0 \\
\bottomrule
\end{tabular}}
\end{table}

Table~\ref{tab:view} details the cross-view results of the main text (Sec.~5.2): back views (closest in geometry to over-shoulder fleet cameras) are easiest, front views hardest. Table~\ref{tab:vis} reports the visibility strata: windows in the eye/face levels, where the upper body and face are well observed, behave like fleet data, whereas head-only and no-body strata are dominated by heavy occlusion and produce several-fold larger errors, quantifying how much partial observability, not just viewpoint, drives difficulty. Table~\ref{tab:head} isolates head-pose forecasting on hardware-grounded labels: with trustworthy targets, learned models do beat head persistence (Transformer 4.19$^\circ$ vs.\ zero-motion 4.80$^\circ$ @4s), in contrast to the vision-derived web labels where annotation noise masks progress, an argument for grounded head pose as a first-class dataset property.

\section{Best-of-$N$ evaluation}
\label{sec:app_bestofn}

\begin{table}[t]
\centering
\caption{Best-of-$N$ evaluation of stochastic models on the fixed stochastic-evaluation subset: minimum ADE/FDE over $N$ samples (23-point set, $10^{-2}$ shoulder widths) and average pairwise diversity (APD). The CVAE's best-of-20 FDE (10.5) beats the best deterministic point forecast (10.9); the DDPM trades accuracy for the largest diversity.}
\label{tab:bestofn}
\small
\setlength{\tabcolsep}{5pt}
\adjustbox{max width=\linewidth}{%
\rowcolors{2}{dmblue3}{white}
\begin{tabular}{@{\extracolsep{\fill}}lcccc}
\toprule
Model & $N$ & minADE $\downarrow$ & minFDE $\downarrow$ & APD \\
\midrule
CVAE (our impl.) & 5 & 10.1 & 11.3 & 1.2 \\
CVAE (our impl.) & 20 & 9.7 & 10.5 & 1.4 \\
DDPM (our impl.) & 5 & 14.0 & 14.2 & 2.0 \\
DDPM (our impl.) & 20 & 13.2 & 12.6 & 2.3 \\
LLM (Llama-3.2-1B + LoRA)~\cite{chen2024motionllm} (our impl.) & 5 & 15.0 & 16.6 & 1.6 \\
\bottomrule
\end{tabular}}
\end{table}

Table~\ref{tab:bestofn} reports minADE/minFDE@$N$ and sample diversity for the stochastic families (autoregressive models sample at $N{=}5$ due to decoding cost and are added with that caveat). Best-of-$N$ is where stochastic forecasting shows its value on this data: sampling 20 CVAE futures and keeping the best beats every deterministic point forecast, confirming that the future is genuinely multimodal and that point metrics alone understate what stochastic models capture.

\section{Probes: intent anticipation and semantic transfer}
\label{sec:app_probes}

If driver motion truly anticipates driving actions, the observed 8\,s of pose should predict maneuver initiations \emph{before they happen}, which also answers, in the affirmative, whether the pre stratum of Protocol A contains real anticipatory signal (P-intent). We label each BATON test window with whether an event of kind $k$ begins within $\tau \in \{1, 2\}$\,s after the observation ends, and train (a) a linear probe on the frozen Transformer forecaster encoder and (b) a small GRU classifier from scratch. Anticipation is real but maneuver-dependent: left turns and hard braking are predicted well above chance (AUPRC $2.8\times$ and $2.9\times$ prevalence at $\tau{=}2$\,s, bootstrap CIs excluding chance), deceleration and acceleration moderately so ($1.5$--$1.9\times$), while lane changes are not separable from chance from pose alone (CIs include prevalence), consistent with mirror-check and posture-shift precursors preceding some maneuvers and not others (full per-kind table with prevalences in Sec.~\ref{sec:app_probes}). The frozen forecaster features match task-specific training, indicating the anticipatory signal is carried by the pose dynamics themselves. Two conclusions follow. Positively: driver motion contains \emph{prospective} information about driving behavior before the vehicle acts, which is what makes the pre stratum of Protocol A meaningful. Negatively, and just as usefully: some behaviors, lane-change precursors above all, are not predictable from skeleton alone, marking the frontier where richer context (scene semantics, gaze, vehicle state) or stochastic prediction is required.

AIDE's 3\,s clips carry behavior and emotion labels, letting us ask what survives in the skeleton-only representation (P-sem). A linear probe on frozen forecaster features classifies \textbf{driver behavior} (7 classes) at 64.7\% accuracy vs.\ a 47.1\% majority-class baseline ($+17.6$\,pp; macro-F1 0.445): actions such as phone use, looking around, and smoking are clearly encoded in body motion. \textbf{Emotion} (5 classes) reaches 64.8\% vs.\ a 60.2\% majority baseline ($+4.6$\,pp): emotion is substantially less linearly decodable from skeletons than behavior. An RGB reference probe on the same splits (Sec.~\ref{sec:app_probes}) quantifies how much of the appearance-borne signal the skeleton format removes: RGB reaches 70.3\% on behavior and 78.3\% on emotion, so the skeleton release retains most of the behavioral signal ($64.7$ vs.\ $70.3$) while attenuating the appearance-dependent emotional cues ($64.8$ vs.\ $78.3$), the privacy--utility trade-off of the release format, made quantitative.

\begin{table}[t]
\centering
\caption{Intent anticipation (P-intent), per maneuver kind on BATON test (10{,}000 windows). AUC of predicting a CAN event maneuver start within $\tau$ seconds after the observation window; probe: linear head on the frozen Transformer forecaster encoder; scratch: small GRU trained end-to-end. AUPRC at $\tau{=}2$\,s is reported with its lift over prevalence (Pos.\,\%); $^\dagger$: bootstrap CI excludes chance. Rare positives make AUPRC the honest severity measure alongside AUC.}
\label{tab:p7}
\small
\setlength{\tabcolsep}{3.5pt}
\adjustbox{max width=\linewidth}{%
\rowcolors{2}{dmblue3}{white}
\begin{tabular}{@{\extracolsep{\fill}}lcccccc}
\toprule
 & \multicolumn{2}{c}{AUC $\tau{=}1$\,s} & \multicolumn{2}{c}{AUC $\tau{=}2$\,s} & AUPRC $\tau{=}2$\,s & \\
Maneuver & Probe & Scratch & Probe & Scratch & (lift) & Pos.\,\% \\
\midrule
Turn left & 0.70 & 0.65 & 0.68 & 0.68 & 0.030 (2.8$\times$)$^\dagger$ & 1.1 \\
Turn right & 0.63 & 0.64 & 0.63 & 0.57 & 0.015 (1.8$\times$)$^\dagger$ & 0.9 \\
Lane change L & 0.60 & 0.59 & 0.59 & 0.61 & 0.005 (1.4$\times$) & 0.4 \\
Lane change R & 0.59 & 0.50 & 0.60 & 0.61 & 0.008 (1.8$\times$) & 0.4 \\
Hard brake & 0.62 & 0.66 & 0.67 & 0.65 & 0.019 (2.9$\times$)$^\dagger$ & 0.7 \\
Decelerate & 0.62 & 0.64 & 0.66 & 0.59 & 0.028 (1.9$\times$)$^\dagger$ & 1.5 \\
Accelerate & 0.60 & 0.56 & 0.63 & 0.58 & 0.050 (1.5$\times$)$^\dagger$ & 3.2 \\
\bottomrule
\end{tabular}}
\end{table}

\paragraph{P-intent setup.} Labels come from the CAN event bank (Sec.~\ref{sec:app_protocol}); a window is positive for kind $k$ if a maneuver start of $k$ falls in $(t_{\text{end}}, t_{\text{end}}{+}\tau]$. We use 40{,}000 training and 10{,}000 test windows, class-balanced loss weights, and report AUC (robust to the 0.1--3\% positive rates); per-kind results are in Table~\ref{tab:p7}. The probe and the from-scratch classifier are comparable overall, indicating the anticipatory signal resides in the pose dynamics rather than in forecaster-specific features.

\paragraph{P-sem setup.} AIDE labels are cleaned before probing: case-duplicate emotion labels (\emph{happiness}/\emph{Happiness}, \emph{weariness}/\emph{Weariness}) are folded, and the singleton vehicle-state label \emph{Forward} is merged into \emph{Forward Moving}. Clips are capped at the benchmark context length; features are the frozen Transformer encoder pooled over the clip; a linear head is trained on train+val sessions and evaluated on the 586 test-session clips. Behavior: 64.7\% accuracy, macro-F1 0.445 over 7 classes (majority 47.1\%). Emotion: 64.8\%, macro-F1 0.442 over 5 classes (majority 60.2\%). Per-class confusion concentrates in \emph{Talking} vs.\ \emph{Normal Driving}, the pair least separable from body motion alone.

\section{Qualitative results}
\label{sec:app_qual}

\begin{figure*}[htbp]
  \centering
  \includegraphics[width=0.99\linewidth,height=0.92\textheight,keepaspectratio]{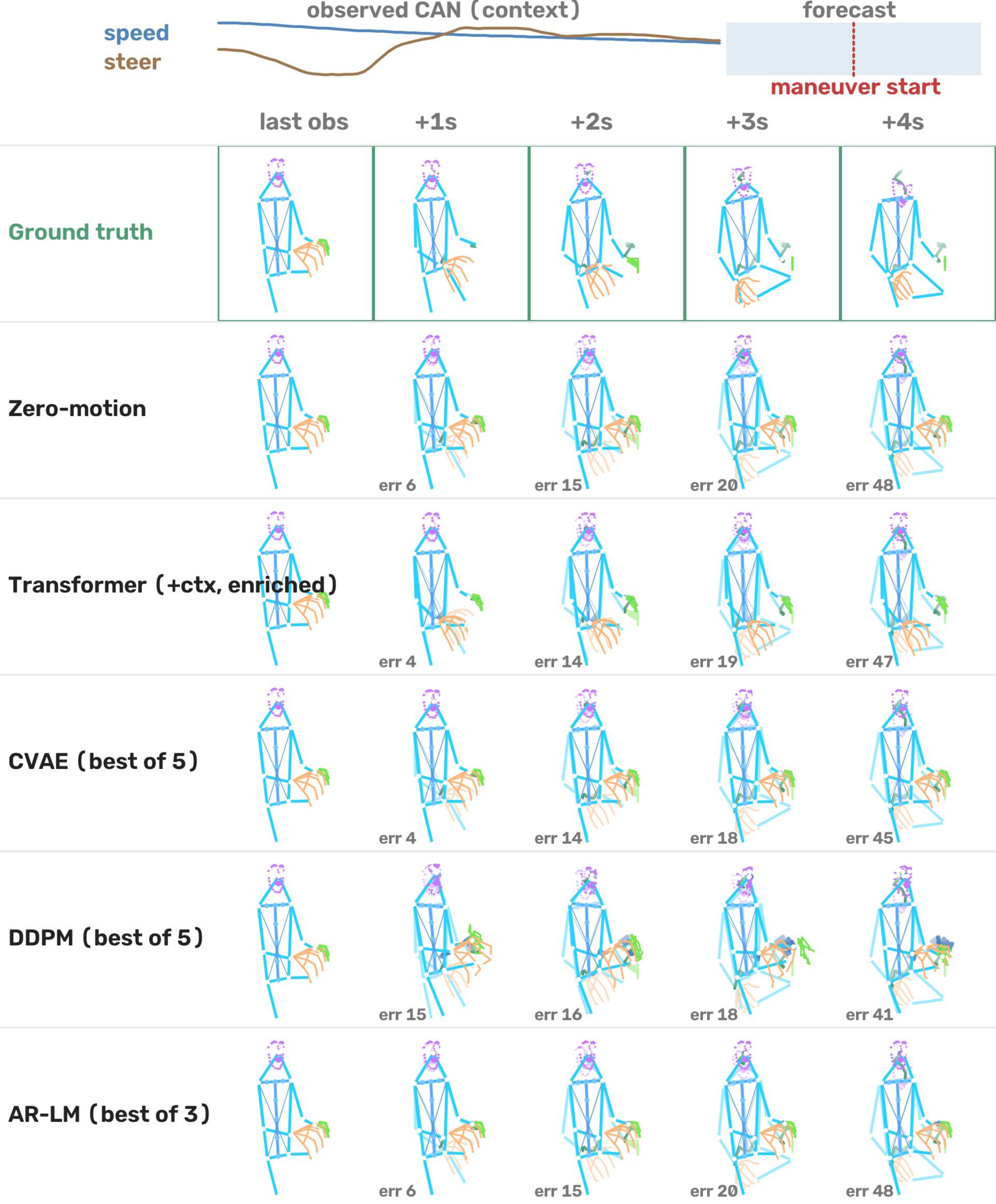}
  \caption{Qualitative forecasts on an anchored pre window. Faint skeletons: ground truth at the same horizon; green/blue trails: ground-truth/predicted nose and wrist motion; numbers: 23-point FDE. The CAN strip is illustrative only.}
  \label{fig:qual}
\end{figure*}

Figure~\ref{fig:qual} makes the board's quantitative patterns visible on one anchored pre window. Ground truth develops a large arm excursion toward the maneuver: the faint reference skeleton pulls away from every prediction as the horizon grows. The failure modes separate by family. Zero-motion is exact at short horizons and collapses at 4\,s. The regression Transformer tracks the torso and head closely but under-commits to the arm, the freeze bias that Part-State F1 penalizes and coordinate error largely forgives. The CVAE's best-of-5 sample behaves like a smoothed regressor. The diffusion model is the opposite: it commits to large hand and arm motion (visible as the most active blue trails) at the cost of skeletal tidiness, matching its profile on the board, weak coordinates, strong motion production. The motion-token AR-LM moves in discrete steps and reaches amplitudes the regressors never produce, consistent with its leading raw Part-State F1. Across all rows, the per-cell FDE numbers are nearly indistinguishable while the produced \emph{behavior} differs sharply, a compact visual argument for reporting both metric axes.

\section{Diagnostics and failure modes}
\label{sec:app_diag}

This section consolidates the negative and diagnostic results referenced from the main text; each is a property of the task worth knowing before designing a model for it.

\begin{table}[t]
\centering
\caption{Diagnostic rows on the Protocol A board (same columns as the main-text board): constant-velocity jitter amplification and the world-model-family baselines (pose-only).}
\label{tab:dyndiag}
\small
\setlength{\tabcolsep}{3pt}
\adjustbox{max width=\linewidth}{%
\rowcolors{3}{white}{dmblue3}
\begin{tabular}{@{\extracolsep{\fill}}lcccccccc}
\toprule
 & \multicolumn{4}{c}{MPJPE@4s $\downarrow$} & \multicolumn{4}{c}{State-F1@2s $\uparrow$} \\
Model & Pre & Post & Stable & Macro & Pre & Post & Stable & Macro \\
\midrule
Constant-velocity & 57.38 & 58.45 & 44.93 & 53.59 & 0.209 & 0.227 & 0.140 & 0.192 \\
\midrule
JEPA\textsuperscript{p}~\cite{assran2023ijepa} & 8.62 & 8.74 & 5.27 & 7.54 & 0.235 & 0.220 & 0.296 & 0.250 \\
JEPA (enriched) & 8.67 & 8.77 & 5.40 & 7.61 & 0.259 & 0.245 & 0.302 & 0.269 \\
Driver-WM$^\dagger$\textsuperscript{p}~\cite{driverwm2026} & 8.86 & 9.08 & 5.34 & 7.76 & 0.196 & 0.180 & 0.269 & 0.215 \\
\bottomrule
\end{tabular}}
\end{table}

\paragraph{Constant velocity amplifies jitter.} Extrapolating instantaneous velocity yields 72.2 MPJPE@4s on the natural distribution and 53.6 macro on the anchored board (Table~\ref{tab:dyndiag}), roughly $7\times$ persistence. Per-frame keypoint jitter (median 0.028 canonical units against a true drift of ${\sim}0.003$) integrates into large phantom displacement over 40 frames; any velocity-based heuristic on pose streams needs temporal smoothing before extrapolation.

\paragraph{A short-horizon world-model recipe approaches a fixed point at long horizons.} The Driver-WM-style adapted baseline matches but does not beat zero-motion on both boards (10.4 @4s natural; macro 7.76 anchored, state-F1 identical to persistence). Its closed-loop latent rollout, designed and validated in the original work for ${\approx}$1.5\,s forecasts, converges to a fixed point over our 40-step horizon. This is a finding about horizon length under our inputs, not about the original system (the main text (Sec.~4.5)); long-horizon driver forecasting is not solved by short-horizon world-model formulations.

\paragraph{Latent-space prediction does not yet convert to pose accuracy.} The JEPA-style predictor lands at persistence-level MPJPE (10.3 @4s natural; 7.54 macro anchored) while producing substantial motion (freeze 0.39) and above-persistence state-F1 (0.250 vs.\ 0.215): its latents rank upcoming motion better than persistence, but the decode from predicted latents loses the geometric precision that direct regression keeps.

\paragraph{Language pretraining does not transfer to motion tokens here.} The Llama-1B-backed model achieves lower token cross-entropy than the task-trained AR-LM but slightly worse decoded geometry (13.0 vs.\ 12.7 @4s), and both trail regression models on coordinates while leading them on forecast-derived torso/arm state-F1 (Table~\ref{tab:state}). Sequence modeling skill transfers; geometric calibration does not.

\paragraph{Privileged dynamics input (upper bound).} An additional configuration that feeds CAN channels as model inputs, excluded from the official track, where vehicle dynamics are construction-time-only, bounds what perfect dynamics context could add: the base-size Transformer with exterior features and privileged CAN reaches 6.61 macro MPJPE@4s (pre 7.63), matching the $6\times$-larger Transformer-L (6.63) trained pose-only. The ${\sim}2\%$ coordinate margin over the official context-conditioned configuration is the headroom available to methods that infer vehicle state from the exterior stream.

\paragraph{Mask-weighted loss ablation.} Removing visibility-mask weighting from the training loss changes MPJPE@4s by less than 0.05 on every architecture tested; the masks matter for evaluation integrity, not for optimization.

\section{Reproducibility}
\label{sec:app_repro}
The release includes the benchmark toolkit (dataset loader, canonicalization, tokenizer configuration, all baseline implementations, the protocol runner with subset hashes, and table generation), the frozen manifest and window lists, per-sequence quality scores, and the CAN event bank, sufficient to reproduce every number in this paper from the released sequence records.

\end{document}